\documentclass[review]{elsarticle}
\usepackage{tikz}
\usetikzlibrary{fit,positioning}
\usepackage{algpseudocode}
\usepackage{algorithmicx}
\usepackage{algorithm}
\usepackage{epsfig}
\usepackage[fleqn]{amsmath}
\usepackage{amsfonts}
\usepackage{pifont}
\usepackage{amsmath}
\usepackage[
    pdfstartview=FitB,
    pdfpagemode=UseOutlines,
    bookmarksnumbered=true,
    colorlinks=true,
    pdftitle={An Algorithm for Learning Shape and Appearance Models without Annotations},
    pdfauthor={Ashburner et al},
    pdfsubject={Shape and Appearance Models},
    pdfkeywords={Machine Learning, Latent Variables, Diffeomorphisms, Geodesic Shooting, Shape Model, Appearance Model}
]{hyperref}
\usepackage{natbib}
\DeclareMathOperator{\Tr}{Tr}
\DeclareMathOperator{\E}{{\mathbb E}}
\DeclareMathOperator*{\argmin}{arg\,min}
\DeclareMathOperator*{\argmax}{arg\,max}
\DeclareMathOperator{\diag}{diag}

\journal{Medical Image Analysis}
\begin{document}
\begin{frontmatter}
\title{An Algorithm for Learning Shape and Appearance Models without Annotations}
\author[mymainaddress]{John Ashburner}
\ead{j.ashburner@ucl.ac.uk}
\author[mymainaddress]{Mikael Brudfors}
\author[mymainaddress]{Kevin Bronik}
\author[mymainaddress]{Ya\"el Balbastre}

\address[mymainaddress]{
Wellcome Centre for Human Neuroimaging\\
UCL Queen Square Institute of Neurology\\
12 Queen Square, London, WC1N 3BG, UK.
}

\begin{abstract}
This paper presents a framework for automatically learning shape and appearance models for medical (and certain other) images.
It is based on the idea that having a more accurate shape and appearance model leads to more accurate image registration, which in turn leads to a more accurate shape and appearance model.
This leads naturally to an iterative scheme, which is based on a probabilistic generative model that is fit using Gauss-Newton updates within an EM-like framework.
It was developed with the aim of enabling distributed privacy-preserving analysis of brain image data, such that shared information (shape and appearance basis functions) may be passed across sites, whereas latent variables that encode individual images remain secure within each site.
These latent variables are proposed as features for privacy-preserving data mining applications.

The approach is demonstrated qualitatively on the KDEF dataset of 2D face images, showing that it can align images that traditionally require shape and appearance models trained using manually annotated data (manually defined landmarks etc.).
It is applied to MNIST dataset of handwritten digits to show its potential for machine learning applications, particularly when training data is limited.
The model is able to handle ``missing data'', which allows it to be cross-validated according to how well it can predict left-out voxels.
The suitability of the derived features for classifying individuals into patient groups was assessed by applying it to a dataset of over 1,900 segmented T1-weighted MR images, which included images from the COBRE and ABIDE datasets.
\end{abstract}

\begin{keyword}
Machine Learning \sep Latent Variables \sep Diffeomorphisms \sep Geodesic Shooting \sep Shape Model \sep Appearance Model
\end{keyword}

\end{frontmatter}


\section{Introduction}
This paper introduces an algorithm for learning a model of shape and appearance variability from a collection of images, without relying on manual annotations.
The shape part of the model concerns modelling variability with diffeomorphic deformations, which is essentially image registration.
In contrast, the appearance part is about accounting for signal variability that is not well described by deformations, and is essentially about adapting a ``template'' to enable more precise registration.

The problem of image registration is often viewed from a Bayesian perspective, whereby the aim is to determine the most probable deformation ($\psi$) given the fixed (${\bf f}$) and moving ($\boldsymbol{\mu}$) images 
\begin{align}
\hat{\boldsymbol\psi} {}={} \argmax_{\psi} \log p(\psi | {\bf f}, \boldsymbol\mu) = \argmax_{\psi} \left( \log p({\bf f} | \psi, \boldsymbol\mu) + \log p(\psi) \right).
\end{align}
In practice, the regularisation term ($\log p(\psi)$) is not usually defined empirically, and simply involves a penalty based on some simple measure of deformation smoothness.
One of the aims of this work is to try to improve on this simple model.
By providing empirically derived priors for the allowable deformations, trained shape models have been shown to exhibit more robust image registration.
An early example is \cite{cootes1992active}, in which control point positions are constrained by their first few modes of variability.
Training this model involved annotating images by manually placing a number of corresponding landmarks, computing the mean and covariance of the collection of landmarks, and then computing the eigenvectors of the covariance \citep{cootes1995active}.
In neuroimaging, shape models have previously been used to increase the robustness of brain image segmentation \citep{babalola2009evaluation,patenaude2011bayesian}.
The current work involves densely parameterised shape models within the diffeomorphic setting, and relates to previous work on diffeomorphic shape models \citep{cootes2008diffeomorphic}, as well as those using more densely parameterised deformations \citep{rueckert2003automatic}.
Recently, \cite{zhang2015bayesian} developed their Principal Geodesic Analysis (PGA) framework for directly computing the main modes of shape variation within a diffeomorphic setting.
The work presented in this paper borrows heavily from that of \cite{zhang2015bayesian}, but extends it to involve a joint model of shape and appearance variability.

In addition to increasing the robustness of image registration tasks, shape models can also provide features that may be used for statistical shape analysis.
This is related to approaches used in geometric morphometrics \citep{adams2004geometric}, where the aim is to understand shape differences among anatomies.
Shape descriptors from the PGA framework have previously been found to be useful features for data mining \citep{zhang2017probabilistic}.

A number of previous works have investigated combining both shape and appearance variability into the same model \citep{cootes1995active,cootes2001active,cootes2001statistical,cootes2008diffeomorphic,belongie2002shape,patenaude2011bayesian}.
These combined shape and appearance models have generally shown good performance in a number of medical imaging challenges \citep{litjens2014evaluation}.
While there is quite a lot written about learning appearance variability alone, the literature on automatically learning both shape and appearance together is fairly limited. 
Earlier approaches required annotated data for training, but there are now some works appearing that have looked into the possibility of using unsupervised or semi-supervised approaches for learning shape and appearance variability.
Examples include \cite{cootes2010computing}, \cite{alabort2014bayesian}, \cite{lindner2015learning} and \cite{stern2016local}.
The current work is about an unsupervised approach, but there is no reason why it could not be made semi-supervised by also incorporating some manually defined landmarks or other features.

This work was undertaken as a task in the Medical Informatics Platform of the EU Human Brain Project (HBP).
The original aim of the Medical Informatics Platform was to develop a distributed knowledge discovery framework that enables data mining without violating patient confidentiality.
The strategy was to involve a horizontally partitioned dataset, where data about different patients is stored in different hospital sites, and only data aggregated over many subjects may leave any site.
While aggregates could be exploited by those with malicious intent in order to extract some information about individual patients, it is generally more difficult to do so, particularly when there are constraints on how much aggregated data may leave each site.
The algorithm presented in this paper can be implemented in a way that does not require patient-specific information to leave a site, and instead only shares aggregates, which reveal less about the individual subjects.
Some leakage of information is inevitable, particularly for sites holding data on only small numbers of individuals, but we leave this as a topic to be addressed elsewhere.

For data mining, in situations where the dimensionality of the data (e.g. number of voxels in an image) is greater than the number of data points (e.g. number of images), it is sometimes possible to make use of the Woodbury matrix identity\footnote{The formula is frequently encountered as $(\boldsymbol{\Sigma}^{-1}+{\bf F}{\bf S}^{-1}{\bf F}^T)^{-1} = \boldsymbol{\Sigma} - \boldsymbol{\Sigma}{\bf F}({\bf S} + {\bf F}^T\boldsymbol{\Sigma}{\bf F})^{-1}{\bf F}^T\boldsymbol{\Sigma}$ (see \url{https://en.wikipedia.org/wiki/Woodbury\_matrix\_identity}). Typically, ${\bf F}$ is an $M \times N$ feature matrix, $\boldsymbol{\Sigma}$ is an $M \times M$ prior covariance and ${\bf S}$ is an $N \times N$ matrix (usually diagonal) that encodes the residual covariance. This is useful when $M \gg N$, as inverses of only $N \times N$ matrices are required. Chapter 2 of \cite{rasmussen2006gaussian} shows how this relates to kernel methods.}, which leads naturally to kernel-based approaches for machine learning.
However, these would require dot-products or differences to be computed between data in different sites, and this would be prohibited by the privacy preserving framework.
Also, the Woodbury identity becomes less useful for extremely large datasets, because of both memory requirements and computational complexity.
Aggregated data may be weighted moments (e.g. $\sum_n r_n$, $\sum_n r_n {\bf z}_n$ or $\sum_n r_n {\bf z}_n {\bf z}_n^T$, where ${\bf z}_n$ is a vector of values for patient $n$, and $r_n$ is a patient-specific weight generated by some rule), which could then be used for clustering or other forms of statistical analysis.
Dimensionality reduction is often needed to enable effective data mining of images, particularly if covariances need to be represented (such as for clustering into patient subgroups using Gaussian mixture models).

Recently, methods such as convolutional neural networks have begun to show a great deal of promise for machine learning tasks \citep{lecun2015deep}, and it seems that much of the previous work from the field of medical imaging will be replaced by new ideas from the machine learning field.
This is partly due to the very large sets of labelled training data now available for certain applications.
When the number of training samples is large compared to the dimensionality of each example, these methods can accurately capture the non-linearities in the training data.
Recent MICCAI and other medical imaging conferences have shown that this is especially true for tasks such as image segmentation, where each voxel is effectively a label.
However, they can be less accurate when the number of training samples is relatively small compared to the complexity of each data point, when other approaches, such as image registration, may be better at encoding the non-linearities \citep{mumford2002pattern}.
Conventional deep learning simply models the output as a nonlinear function of the input.
In contrast, generative approaches to machine learning would involve constructing probabilistic models of the input, such that random samples could be drawn that are as similar as possible to possible future examples of real input data.
A recent incarnation is the family of generative adversarial networks \citep{goodfellow2014generative}, where the general aim is to learn a model that can generate samples that are indistinguishable from real data.
Accurate generative models of data have many potential applications, which include outlier detection and augmenting training data for deep learning.
There is increasing interest in the use of generative approaches for machine learning, partly because they can be extended to work in a semi-supervised way.
This enables unlabelled training data to contribute towards the model, potentially allowing more complex models to be learned from fewer labelled examples.
A simple example of semi-supervised learning would be the approach of \cite{blaiotta2018generative}, which used both labelled and unlabelled images for generating tissue probability maps.

Another motivation for generative modelling approaches is to enable missing data to be dealt with.
Brain images -- particularly hospital brain images -- often have different fields of view from each other, with parts of the brain missing from some of the scans.
Many machine learning approaches do not work well in the presence of missing data, so imputing missing information is an implicit part of the framework presented in this paper.

This work proposes a solution based on learning a form of shape and appearance model.
The overall aim is to capture as much anatomical variability as possible using a relatively small number of latent variables.
In addition to 3D brain image data, a number of other types of images will be used to illustrate other aspects of the very general framework that we present.
Often, it can be difficult to understand the behaviour of an algorithm when it is only applied to 3D volumes, particularly when space for figures in papers is limited.
For this reason, the work is also applied to a number of 2D images, ranging from the tiny images that make up the MNIST dataset, to some images of faces, along with some single slices through brain images.


\section{Methods}
The proposed framework builds heavily on the principal geodesic analysis model of \cite{zhang2015bayesian}.
Modifications involve extending the framework to use a Gauss-Newton optimisation strategy and incorporating an appearance model.
The Gauss-Newton strategy is intended to be more suited to the distributed computing framework of the HBP, although a streaming method using stochastic gradient descent could also have been used.

The basic idea is that both shape and appearance may be modelled by linear combinations of spatial basis functions, and the objective is to automatically learn the best set of basis functions and latent variables from some collection of images.
This is essentially a form of factorisation of the data.
Each of the $N$ images will be denoted by ${\bf f}_n \in \mathcal{R}^M$, where $M$ is the number of pixels/voxels in an image, $1\le n\le N$, and the entire collection of images by ${\bf F}$.
An \emph{appearance model} for the $n$th image is constructed from a linear combination of basis functions, such that
\begin{align}
{\bf a}_n {}={} \boldsymbol\mu + {\bf W}^a {\bf z}_n.
\end{align}
Here, ${\bf W}^a$ is a matrix containing $K$ appearance basis functions, and ${\bf z}_n$ is a vector of $K$ latent variables for the $n$th image.
The vector $\boldsymbol\mu$ is a mean image, with the same dimensions as a column of ${\bf W}^a$.
The \emph{shape model} is encoded similarly, with
\begin{align}
{\bf v}_n {}={} {\bf W}^v {\bf z}_n.
\end{align}
The approach involves warping ${\bf a}_n$ to match ${\bf f}_n$, using a diffeomorphic deformation parameterised by ${\bf v}_n$.
The diffeomorphism ($\psi_n$) is constructed from ${\bf v}_n$ by a procedure known as ``geodesic shooting'', which is outlined in Section \ref{Sec:Shape}.
From a probabilistic perspective, the likelihood can be summarised by
\begin{align}
p({\bf f}_n |  {\bf z}_n, \boldsymbol\mu, {\bf W}^a, {\bf W}^v) {}={} p({\bf f}_n | {\bf a}_n(\psi_n)). \label{Eqn:Appearance}
\end{align}
Different forms of appearance model are presented in Section \ref{Sec:Appearance}, but for convenience, we use the generic definition
\begin{align}
J({\bf f}_n, {\bf z}_n, \boldsymbol\mu, {\bf W}^a, {\bf W}^v) {}={} -\ln p({\bf f}_n |  {\bf z}_n, \boldsymbol\mu, {\bf W}^a, {\bf W}^v).
\end{align}

In practice, a small amount of regularisation is imposed on the mean ($\boldsymbol\mu$) by assuming it is drawn from a multivariate Gaussian distribution of precision ${\bf L}^{\mu}$
\begin{align}
p(\boldsymbol\mu) {}={} & \mathcal{N}(\boldsymbol\mu | {\bf 0}, {({\bf L}^{\mu})}^{-1}).       \label{Eqn:AppReg1}
\end{align}
The precision matrix is only involved conceptually.
Because the matrix is circulant (a special kind of Toeplitz matrix where data are assumed to wrap around at the boundaries), matrix multiplication and division may be effected by 3D convolutions.
Possible forms for ${\bf L}^{\mu}$ are described in Section \ref{Sec:AppReg}.

A weighted average of two strategies for regularising the basis functions (${\bf W}^a$ and ${\bf W}^v$) and latent variables (${\bf z}_n$) is used, which are:
\begin{enumerate}
\item
The first strategy involves separate priors on the basis functions, and on the latent variables.
Each of the basis functions is assumed to be drawn from zero-mean highly multivariate Gaussian, parameterised by very large and sparse precision matrices.
Possible forms of the matrices for regularising shape (${\bf L}^v$) are described in Section \ref{Sec:ShapeReg}, whereas those for appearance (${\bf L}^a$) are described in Section \ref{Sec:AppReg}.
Priors for the basis functions are
\begin{align}
p({\bf W}^v) {}={} & \prod_{k=1}^{K^v} \mathcal{N}({\bf w}^v_k | {\bf 0}, {(N {\bf L}^v)}^{-1}), \label{Eqn:ShapeReg} \\
p({\bf W}^a) {}={} & \prod_{k=1}^{K^a} \mathcal{N}({\bf w}^a_k | {\bf 0}, {(N {\bf L}^a)}^{-1}). \label{Eqn:AppReg2}
\end{align}
The latent variables (${\bf Z}$) are assumed to be drawn from zero-mean multivariate Gaussian distributions, parameterised by a precision matrix (${\bf A}$) that is derived from the data\footnote{Note that the precision matrix ${\bf A}$ should not be confused with the variables ${\bf a}_n$, which were introduced earlier. Hopefully, the context in which they are used should be enough to prevent any confusion.}.
\begin{align}
p({\bf z}_n | {\bf A}) {}={} \mathcal{N}({\bf z}_n | {\bf 0}, {\bf A}^{-1}).
\end{align}
Note that precision matrices for the basis functions, ${\bf L}^v$ and ${\bf L}^a$, are scaled by $N$.
The reason for this is that it keeps the variance of the latent variables within a similar range, irrespective of how many images are involved, which makes it easier to define priors for them.

The simplest approach for working with the ${\bf A}$ matrix is to make a point estimate based on the distribution of the latent variables.
This approach allows the relevance of each of the basis functions (columns of ${\bf W}^a$ and ${\bf W}^v$) to be automatically determined.
If they are not needed, some of the elements of the latent variables (i.e. rows of ${\bf Z}$) are forced to zero \citep{zhang2015bayesian}.
In practice, this could lead to prematurely sparse solutions that may be difficult to recover from.
Instead, sparsity is avoided by assuming that matrix ${\bf A}$ is drawn from a Wishart distribution.
\begin{align}
p({\bf A}) {}={} & \mathcal{W}_K({\bf A} | \boldsymbol\Lambda_0, \nu_0)\cr
           {}={} & \frac{|{\bf A}|^{(\nu_0 - K - 1)/2} \exp (-\tfrac{1}{2}\Tr(\boldsymbol\Lambda_0^{-1} {\bf A}))}{2^{(\nu_0 K)/2} |\boldsymbol\Lambda_0|^{\nu_0/2} \Gamma_K\left(\tfrac{\nu_0}{2}\right)},
\end{align}
where $\Gamma_K$ is the multivariate gamma function.
This prior can be made as uninformative as possible by using $\nu_0 = K$ and $\Lambda_0 = {\bf I}/\nu_0$, where ${\bf I}$ is an identity matrix.
In general, $\boldsymbol\Lambda_0$ should be a positive definite symmetric matrix, with $\nu_0 \ge K$ so that the distribution can be normalised.

\item
The second strategy is similar to that of \cite{zhang2015bayesian}, and is a pragmatic solution to ensuring that enough regularisation is used.
\begin{align}
\ln p({\bf Z},{\bf W}^a,{\bf W}^v) {}={} -\tfrac{1}{2} \Tr({\bf Z}{\bf Z}^T (({\bf W}^a)^T {\bf L}^a {\bf W}^a + ({\bf W}^v)^T {\bf L}^v {\bf W}^v)) + \text{const}
\end{align}
This strategy imposes smoothness on the reconstructions by assuming penalties based on $\ln \mathcal{N}({\bf W}^a {\bf z}_n | {\bf 0}, {\bf L}^a)$ and $\ln \mathcal{N}({\bf W}^v {\bf z}_n | {\bf 0}, {\bf L}^v)$, in a similar way to more conventional regularisation approaches.
\end{enumerate}

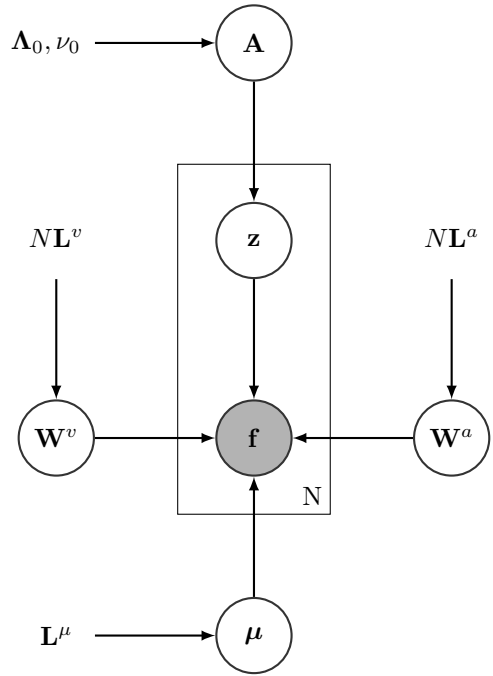
\begin{figure}
\centering
\begin{tikzpicture}
\tikzstyle{main}=[circle, minimum size = 10mm, thick, draw =black!80, node distance = 16mm]
\tikzstyle{connect}=[-latex, thick]
\tikzstyle{box}=[rectangle, draw=black!100]
  \node[main, fill = black!30 ] (F)                 {${\bf f}$};
  \node[main, fill = white!100] (Z)   [above=of F]  {${\bf z}$};
  \node[main, fill = white!100] (A)   [above=of Z]  {${\bf A}$};
  \node[main, fill = white!100] (Wa)  [right=of F]  {${\bf W}^a$};
  \node[main, fill = white!100] (Wv)  [ left=of F]  {${\bf W}^v$};
  \node[main, fill = white!100] (mu)  [below=of F]  {${\boldsymbol{\mu}}$};
  \node[main, draw = white!100] (Lv)  [above=of Wv] {$N {\bf L}^v$};
  \node[main, draw = white!100] (La)  [above=of Wa] {$N {\bf L}^a$};
  \node[main, draw = white!100] (Lmu) [ left=of mu] {${\bf L}^{\mu}$};
  \node[main, draw = white!100] (Lambda0) [ left=of A] {$\boldsymbol\Lambda_0, \nu_0$};

  \path (Z)   edge [connect] (F)
        (Wv)  edge [connect] (F)
        (Wa)  edge [connect] (F)
        (mu)  edge [connect] (F)
        (Lv)  edge [connect] (Wv)
        (La)  edge [connect] (Wa)
        (Lmu) edge [connect] (mu)
        (Lambda0) edge [connect] (A)
        (A)   edge [connect] (Z);
  \node[rectangle, inner sep=0mm, fit= (F) (Z), label=south east:N] {};
  \node[rectangle, inner sep=5mm,draw=black!100, fit= (F) (Z)] {};
\end{tikzpicture}
\caption{A graphical representation of the model (showing only the 1st strategy). Gray circles indicate observed data, whereas white circles indicate variables that are either estimated (${\bf W}^v$, ${\bf W}^a$, $\boldsymbol\mu$ and ${\bf z}$) or marginalised out (${\bf A}$). The plate indicates replication over all images. \label{Fig:graphic}}
\end{figure}

The relative weighting of the two strategies is by $\lambda_1$ and $\lambda_2$.
Better results are typically achieved when hyper-parameters are specified so that they sum to a value greater than 1.
When everything is combined (see Fig. \ref{Fig:graphic}), the following joint log-probability is obtained
\begin{align}
\ln p({\bf F}, & \boldsymbol\mu, {\bf W}^a, {\bf W}^v, {\bf A}, {\bf Z})\cr
{}={}
& - \sum_{n=1}^N J({\bf f}_n, {\bf z}_n, \boldsymbol\mu, {\bf W}^a, {\bf W}^v) - \tfrac{1}{2} \boldsymbol\mu^T {\bf L}^\mu \boldsymbol\mu \cr
& - \tfrac{\lambda_1 N}{2} \left( \Tr(({\bf W}^a)^T {\bf L}^a {\bf W}^a) + \Tr(({\bf W}^v)^T {\bf L}^v {\bf W}^v) \right)\cr
& + \tfrac{\lambda_1}{2} \left((N + \nu_0 - K - 1) \ln |{\bf A}| - \Tr(({\bf Z}{\bf Z}^T + \boldsymbol\Lambda_0^{-1}) {\bf A}) \right)\cr
& - \tfrac{\lambda_2}{2} \Tr({\bf Z}{\bf Z}^T (({\bf W}^a)^T {\bf L}^a {\bf W}^a + ({\bf W}^v)^T {\bf L}^v {\bf W}^v)) + \text{const}. \label{Eqn:Objective}
\end{align}

Fitting the model is described in Appendix \ref{Sec:Algorithm}.
Ideally, the procedure would compute distributions for all variables, such that uncertainty was dealt with optimally.
Unfortunately, this is computationally impractical for the size of the datasets involved.  Instead, only point estimates are made for the latent variables ($\hat{\bf z}_n$) and various parameters ($\hat{\boldsymbol\mu}$, ${\bf W}^a$, ${\bf W}^v$, and sometimes $\hat\sigma^2$), apart from ${\bf A}$, which is inferred within a variational Bayesian framework.

The approach also allows shapes and appearances to be modelled separately, by having some of the latent variables control appearance, and others control shape.
This may be denoted by
\begin{align}
{\bf a}_n {}={} & \boldsymbol\mu + \sum_{k=1}^{K^a} {\bf w}^a_k z_{k,n},\\
{\bf v}_n {}={} & \sum_{k=1}^{K^v} {\bf w}^v_k z_{K^a + k, n}.
\end{align}
For simplicity, only the form where each latent variable controls both shape and appearance is described in detail.
This is the form used in active appearance models \citep{cootes2001active}.
Note however, that in the form where shape and appearance are controlled by separate latent variables, the precision matrix ${\bf A}$ still encodes covariance between the two types of variables.  This means that latent variables controlling either shape or appearance are not estimated completely independently.

\subsection{Shape Model \label{Sec:Shape}}
Relative shapes are encoded by diffeomorphisms, which are each parameterised by a set of initial conditions (${\bf v}_n = {\bf W}^v {\bf z}_n$).
This subsection mixes both discrete and continuous representations of the same objects, so some notation is now introduced.
For the discrete case, where a velocity field is treated as a vector, it are denoted by ${\bf v}_n$.
Alternatively, the same object may be treated as a continuous 3D vector field, where it is denoted by $v_n$.

In addition, deformations may be treated as discrete or continuous.
Within the continuous setting, warping an image by a diffeomorphism $\psi$ may be denoted by $a' = a(\psi)$.
In the discrete setting, this resampling may be conceptualised as a matrix multiplication, where a very large sparse matrix $\boldsymbol\Psi$ encodes the same deformation (and associated trilinear interpolation), such that ${\bf a}' = \boldsymbol\Psi {\bf a}$.
The transpose of this matrix can be used to perform a push-forward operation, which is frequently used in this work and which we denote by ${\bf f}' = \boldsymbol\Psi^T {\bf f}$.

Diffeomorphisms are created within the Large-Deformation Diffeomorphic Metric Mapping (LDDMM) framework \citep{beg2005computing}, which allows image registration to produce smooth, invertible one-to-one mappings.
A simplification of the basic idea is that large deformation diffeomorphisms may be computed by composing together a series of much smaller deformations.
Providing the constituent deformations are one-to-one, then the result of their composition should also be one-to-one.
The original LDDMM implementation involved a variational optimisation, which involves estimating a series of velocity fields ($v_t$) at different ``time-points'', $t$.
An alternative to the variational approach involves estimating only an initial velocity ($v_0$), and deriving velocities at subsequent time points by a procedure known as ``geodesic shooting'' \citep{miller2006geodesic}.
Algorithm \ref{Alg:shoot} in the Appendix presents a simple algorithm for geodesic shooting, which generates a deformation $\psi$ from an initial velocity field $v$.

Another important feature is that when image registration is formulated as a generative model of the individual images, the diffeomorphic framework can be set up such that the parametrisation of the deformations (i.e., the initial velocity fields) are all in alignment with each other in the space of the template.
This makes them more useful for various forms of multivariate models of shape variability as they are better encoded by factorisation models.

\subsubsection{Differential operator for shape model \label{Sec:ShapeReg}} 
Our implementation of geodesic shooting uses Fast Fourier Transform (FFT) methods to obtain the Green's function \citep{bro1996fast}.
Because FFTs are used throughout the work, the boundary conditions for the velocity fields are assumed to be periodic.
The precision matrix used in Eqn. \ref{Eqn:ShapeReg} has the form
\begin{align}
{\bf v}^T {\bf L}^v {\bf v}
 {}={} & \int_{x\in \Omega} \left(\omega^v_0 \| v(x) \|^2 + \omega^v_1 \| \nabla v(x)\|^2 + \omega^v_2 \| \nabla^2 v(x)\|^2 \right) dx\cr
 + & \int_{x\in \Omega} \left(\frac{\omega^v_3}{4} \| Dv(x) + (Dv(x))^T\|_F^2  + \omega^v_4 \Tr(Dv(x))^2 
   \right) dx
\end{align}
where $|\cdot|_F$ denotes the Frobenius norm (the square root of the sum of squares of the matrix elements).
The above integral is defined in Sobolev space, which is a weighted Hilbert space where spatial derivatives, up to a certain degree, are accounted for.

Five hyper-parameters are involved:
\begin{itemize}
\item{$\omega^v_0$ controls absolute displacements, and is typically set to be a very small value.}
\item{$\omega^v_1$ controls stretching, shearing and rotation.}
\item{$\omega^v_2$ controls bending energy.  This ensures that the resulting velocity fields have smooth spatial derivatives.}
\item{$\omega^v_3$ controls stretching and shearing (but not rotation).}
\item{$\omega^v_4$ controls the divergence, which in turn determines the amount of volumetric expansion and contraction.}
\end{itemize}
Most of the regularisation in this work was typically based on a combination of the linear-elasticity ($\omega^v_3$ and $\omega^v_4$) and bending energy ($\omega^v_2$) penalties.
The differential operator involved very little penalty against absolute displacements ($\omega^v_0$), although some was necessary to ensure that the Green's function could be computed, while still allowing the deformations to incorporate uniform translations.
The types of differential operators used in \cite{beg2005computing} would be constructed from $\omega^v_0$, $\omega^v_1$ and $\omega^v_2$, whereas those used in \cite{christensen1996deformable} were largely based on \emph{Lam\'e's constants} $\omega^v_3$ and $\omega^v_4$.
The latter parametrisation allows the stresses and strains of deformations to be formulated in terms of, for example, Poisson's ratio and Young's modulus\footnote{See \url{https://en.wikipedia.org/wiki/Linear_elasticity}.}.

\subsection{Appearance Models \label{Sec:Appearance}}
A number of different choices for the appearance model are available for Eqn. \ref{Eqn:Appearance}, each suitable for modelling different types of image data.
These models are based on $p({\bf f}_n | {\bf a}'_n)$, which leads to an ``energy'' term ($J$) that drives the model fitting and is assumed to be independent across voxels
\begin{align}
{\bf a}'_n & {}={} \boldsymbol{\Psi}_n (\boldsymbol\mu + {\bf W}^a {\bf z}_n)\\
J({\bf a}'_n) & {}={} -\ln p({\bf f}_n |  {\bf a}'_n) = -\sum_{m=1}^M \ln p(f_{mn} | a'_{mn}).
\end{align}
Because the approach is generative, missing data can be handled by simply ignoring those voxels where there is no information.
By doing this, they do not contribute towards the objective function and play no role in driving the model fitting.
A number of different ``energy'' functions have been implemented for modelling different types of data.
These are listed next.

\subsubsection{Gaussian noise model}
Mean-squares difference is a widely used objective functions for image matching, which is based on the assumption of stationary Gaussian noise.
For an image consisting of $M$ pixels or voxels, the function would be
\begin{align}
-J_{L_2}({\bf a}') {}={} \ln p({\bf f} | {\bf a}', \sigma^2) = -\tfrac{M}{2} \ln(2\pi) - \tfrac{M}{2} \ln \sigma^2 - \tfrac{1}{2\sigma^2}||{\bf f}-{\bf a}'||_2^2,
\end{align}
where $||\cdot||_2$ denotes the Euclidean norm.
The simplest approach to compute $\sigma^2$ is to make a maximum likelihood estimate from the variance by
\begin{align}
\hat{\sigma^2} {}={} \tfrac{1}{MN} \sum_{n=1}^N ||{\bf f}_n-{\bf a}'_n||_2^2 .
\end{align}

\subsubsection{Logistic function with Bernoulli noise model}
When working with binary images, such as single tissue type maps having voxels of zeros and ones (or values very close to zero or one), it may be better to work under the assumption that voxels are drawn from a Bernoulli distribution, which is a special case of the binomial distribution.
For a single voxel,
\begin{align}
P(f | s) {}={} s^{f}(1-s)^{1-f}.
\end{align}

The range $0<s<1$ must be satisfied, which can be achieved using a logistic sigmoid function
\begin{align}
s(a') {}={} \frac{1}{1+\exp(-a')}.
\end{align}

Putting these together gives
\begin{align}
P(f | a') & {}={} \left(\frac{1}{1+\exp(-a')} \right)^{f} \left(1-\frac{1}{1+\exp(-a')}\right)^{1-f}\cr
& {}={} \exp(a' f) s(-a').
\end{align}

This gives the matching function
\begin{align}
-J_{Bern}({\bf a}') {}={} \ln P({\bf f} | {\bf a}') = \sum_{m=1}^M \left(f_m a'_m + \ln s(-a'_m)\right).
\end{align}

\subsubsection{Softmax function with categorical noise model}
If there are several binary maps to align simultaneously, for example maps of grey matter, white matter and background, then a categorical noise model is appropriate.
A categorical distribution is a generalisation of the Bernoulli distribution, and also a special case of the multinomial distribution.
The probability of a vector ${\bf f}$ of length $C$, such that $f_c \in \{0,1\}$ and $\sum_{c=1}^C f_c = 1$, is given by 
\begin{align}
P({\bf f} | {\bf s}) {}={} \prod_{c=1}^C s_c^{f_c},
\end{align}
where $s_c>0$ and $\sum_{c=1}^C s_c = 1$.
The constraints on ${\bf s}$ can be enforced by using a softmax function.
\begin{align}
s_c({\bf a}') {}={} \frac{\exp a'_c}{\sum_{c=1}^C \exp a'_c}
\end{align}
Using the ``log-sum-exp trick'', numerical overflow or underflow can be prevented by first subtracting the maximum of ${\bf a}$, so
\begin{align}
s_c({\bf a}') {}={} \frac{\exp(a'_c-a^*)}{\sum_{c=1}^C \exp(a'_c-a^*)}
\text{, where } a^* {}={} \max\{a'_{1},\hdots,a'_{C}\}
\end{align}

Noting that each image is now a matrix of $M$ voxels and $C$ classes, the objective function can then be computed as
\begin{align}
-J_{cat}({\bf A}') & {}={} \ln P({\bf F} | {\bf A}')\cr
& {}={} \sum_{m=1}^M \left(\sum_{c=1}^C a'_{mc} f_{mc} - a^* - \log\left(\sum_{c=1}^C \exp(a'_{mc}-a_m^*) \right) \right)
\end{align}

\subsubsection{Differential operator for appearance model \label{Sec:AppReg}}
Regularisation is required for the appearance variability, as it helps to prevent the appearance model from absorbing too much of the variance, at the expense of the shape model.
This differential operator (again based on a Sobolev space) is used in Eqns. \ref{Eqn:AppReg1} and \ref{Eqn:AppReg2}, and controlled by three hyper-parameters.
\begin{align}
{\bf a}^T {\bf L}^a {\bf a}  
 {}={} \int_{x \in \Omega} \left(\omega^a_0 \| a(x) \|^2 + \omega^a_1 \| \nabla a(x)\|^2 + \omega^a_2 \| \nabla^2 a(x)\|^2 \right) dx
 \end{align}
The extent to which intensity differences may influence apparent spatial deformations was briefly discussed in \cite{evans1995commentary}.
Automatically disambiguating between what should be modelled as a shape change, and what should be treated as as appearance change, is still a largely unresolved area.
The approach taken in this work is simply to treat the construct as a model of the data, and to assess it according to how well it describes and predicts the observations, rather than how well it can separately estimate shape information versus appearance information.


\section{Results}
To show the general applicability of the approach, evaluations were performed with a number of datasets of varying characteristics.
Our implementation\footnote{Available from \url{https://github.com/WTCN-computational-anatomy-group/Shape-Appearance-Model}.} is written in a mixture of MATLAB and C code (MATLAB ``mex'' files for the computationally expensive parts).

\subsection{Qualitative 2D experiments with faces}
After years of exposure to faces, most people can identify whether an image of a face is plausible or not, so images of human faces provide a good qualitative test of how well the algorithm can model biological variability.
Recent evidence \citep{chang2017code} suggests that the primate brain might even use some form of shape and appearance model for encoding faces, which might indicate that the type of model used in this work could be quite effective.

The straight on views from the Karolinska Directed Emotional Faces (KDEF) data-set \citep{lundqvist1998karolinska} were used to make a visual assessment of how well the algorithm performs.
This data-set consisted of photographs of 70 participants, holding seven different facial expressions, which was repeated twice.  Some of the images were not usable, so the final dataset consisted of 932 colour images, which were downsampled to a size of $282 \times 382$.
The original intensities were in the range of 0 to 255, but these values were re-scaled by $1/255$.

\begin{figure}
\begin{center}
Original KDEF images.\\
\epsfig{file=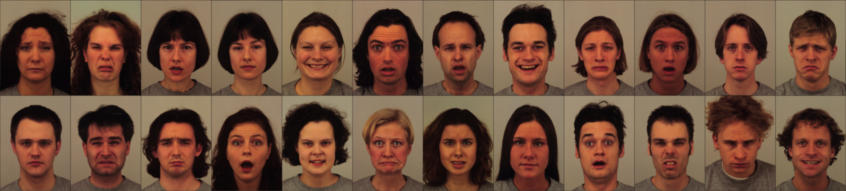,width=\textwidth}\\
Full shape and appearance fit.\\
\epsfig{file=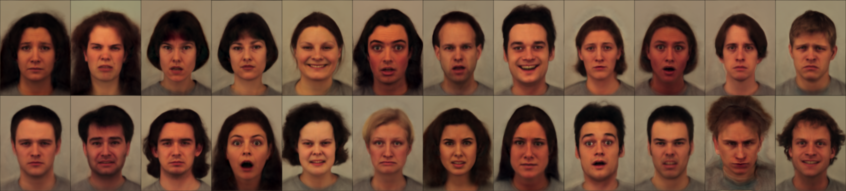,width=\textwidth}\\
Shape fit only.\\
\epsfig{file=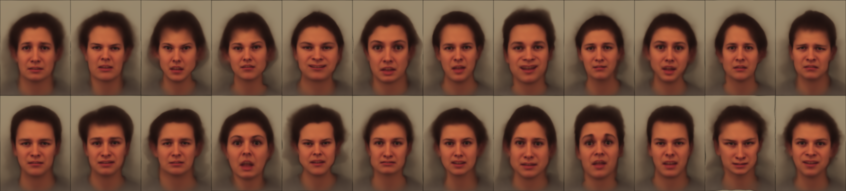,width=\textwidth}\\
Appearance fit only.\\
\epsfig{file=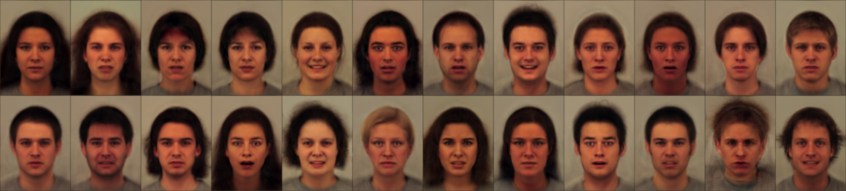,width=\textwidth}\\
\end{center}
\caption{
Shape and appearance fit shown for a randomly selected sample of the KDEF face images. \label{Fig:KDEF_full}}
\end{figure}

\begin{figure}
\begin{center}
\epsfig{file=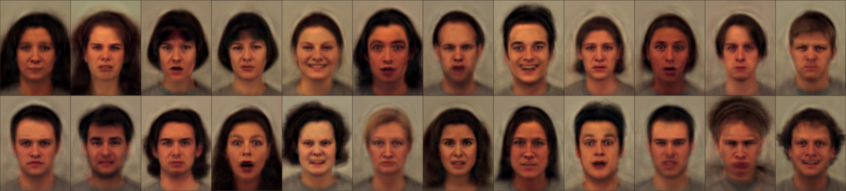,width=\textwidth}\\
\vspace{2mm}
\epsfig{file=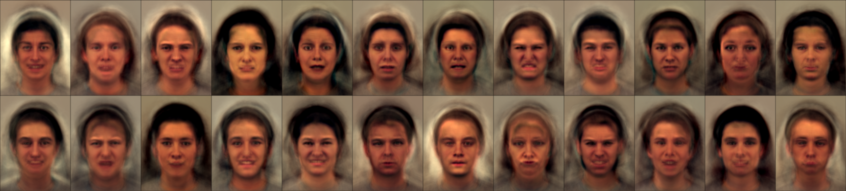,width=\textwidth}\\
\end{center}
\caption{
Fits using a simple 64-mode principal component analysis model are shown above (cf. Fig. \ref{Fig:KDEF_full}), and random faces generated from the same PCA model are shown below (cf. Fig. \ref{Fig:ColourFaces_random}).
\label{Fig:ColourFaces_random_pca}}
\end{figure}

A 64 eigenmode model was used ($K=64$), which assumed Gaussian noise.
Model fitting (i.e., learning the shape and appearance basis functions, etc.) was run for 20 iterations, with $\nu_0 = 1000$, $\lambda = [15.2 \ 0.8]$, $\omega^a = [4 \ 512 \ 64]$, $\omega^{\mu} = N [10^{-4} \ 0.1 \ 0.1]$ and $\omega^v = [10^{-3} \ 0 \ 16 \ 1 \ 1]$.
It was fit to the entire field of view of the images, rather than focusing only on the faces, and some of the resulting fits are shown in Fig. \ref{Fig:KDEF_full}.
The first set of images are a random selection of the original data, with the full shape and appearance model fits shown immediately below.
As can be seen, the fit is reasonably good - especially given that only 64 modes of variability were used, and that these have to account for a lot of variability of hair etc.
Below these are the shape model fits, generated by warping the mean according to the estimated deformations ($\boldsymbol\mu(\psi_n)$).
The appearance fits are shown at the bottom (${\bf a}_n$ from Eqn. \ref{Eqn:Appearance}).
Ideally, these reconstructions of appearance should be in perfect alignment with each other, which is not quite achieved in certain parts of the images.
In particular, the thickness of the neck varies according to whether or not the people in the images have short or long hair.
When looked at separately, the shape and appearance parts of the model do not behave quite so well, but when combined, they give quite a good fit.
Fig. \ref{Fig:ColourFaces_random_pca} shows a simple 64-mode principal component analysis (PCA) fit to the same data, which clearly does not capture variability quite as well as the shape and appearance model.

\begin{figure}
\begin{center}
\epsfig{file=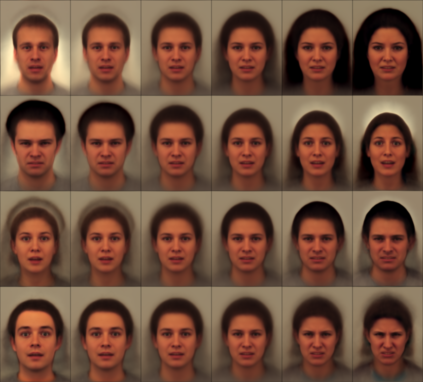,width=.32\textwidth}
\epsfig{file=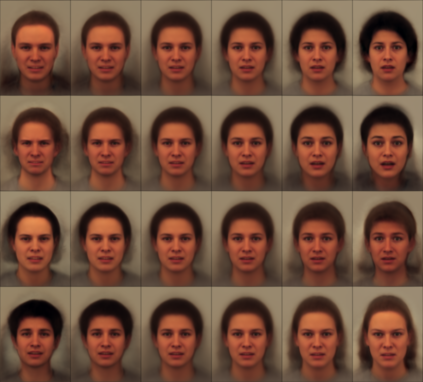,width=.32\textwidth}
\epsfig{file=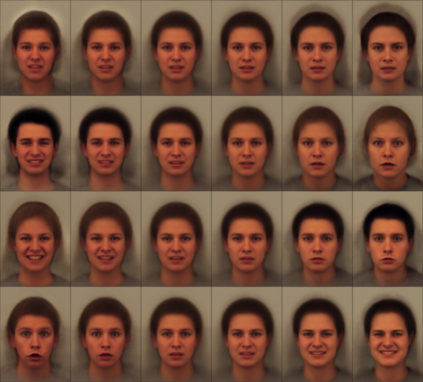,width=.32\textwidth}
\end{center}
\caption{
The first 12 modes of variability, shown (left to right) at -5, -3, -1, 1, 3 and 5 standard deviations (s.d.).\label{Fig:KDEFmodes}}
\end{figure}

The first 12 modes of variability are illustrated in Fig. \ref{Fig:KDEFmodes}.
Much of the variance to be modelled is not actually part of the face, which can be seen in how the hairstyles of the subjects drive the first few modes.
Better fits would have been achieved if only the faces themselves had been modelled, but the main aim here was to illustrate the behaviour of the approach in more challenging situations.
Masking out parts of the data from model fitting is possible, as one of the objectives of the work was to be able to handle missing data (see Section \ref{Sec:Missing}).

For these examples, there should really have been a distinction between inter-subject variability and intra-subject variability, using some form of hierarchical model for the latent variables.
This type of hierarchical mixed-effects model is widely used for analysing multi-subject data within the neuroimaging field \citep{friston2002classical}, and a number of works have applied mixed effects modeling to image registration \citep{datar2012mixed,allassonniere2015bayesian}.

Spatial correlations in the images mean the assumptions about the independence of the noise across neighbouring voxels to be less valid, which causes problems for the simple i.i.d. noise model used here.
This is why the regularisation needed to be weighted more heavily relative to the image matching term.
This is related to the ``virtual decimation'' \citep{groves2011linked} approach of down-weighting the matching term by a correction factor that accounts for the number of independent observations.
Numbers of independent observations could be derived using random field theory \citep{worsley1995tests}, using gradients of residuals to estimate image smoothness in terms of the full-width half maximum of a Gaussian.

\subsubsection{Simulating faces}
\begin{figure}
\begin{center}
\epsfig{file=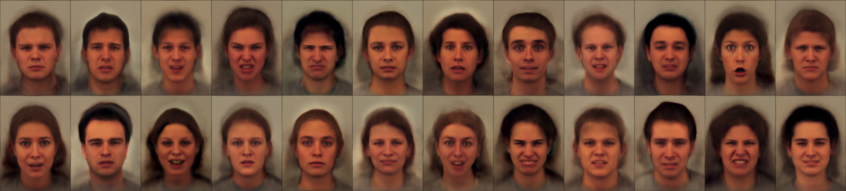,width=\textwidth}\\
\vspace{2mm}
\epsfig{file=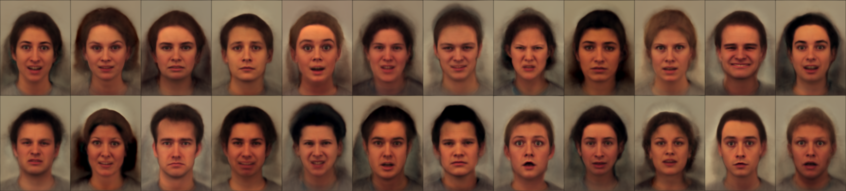,width=\textwidth}
\end{center}
\caption{
Random faces generated from the shape and appearance model.  The lower set of faces were generated with the same latent variables as those shown in the upper set, except the values were multiplied by -1 and thus show a sort of ``opposite'' face. For example, if a face in the top set has a wide open mouth, then the mouth should be tightly closed in the corresponding image of the bottom set. \label{Fig:ColourFaces_random}}
\end{figure}

Once the model is learned, it becomes possible to generate random faces from the estimated distribution.
This involves drawing a random vector of latent variables ${\bf z} \sim \mathcal{N}({\bf 0},\hat{\bf A}^{-1})$, and using these to reconstruct a face.
Fig. \ref{Fig:ColourFaces_random} shows two sets of randomly generated faces, where the lower set used the same latent variables as the upper set, except that they were multiplied by -1.
Although some of the random faces are not entirely plausible, they are much more realistic than faces generated from a simple 64-mode PCA model (shown in Fig. \ref{Fig:ColourFaces_random_pca}).

\subsubsection{Vector arithmetic}
\begin{figure}
\begin{center}
\epsfig{file=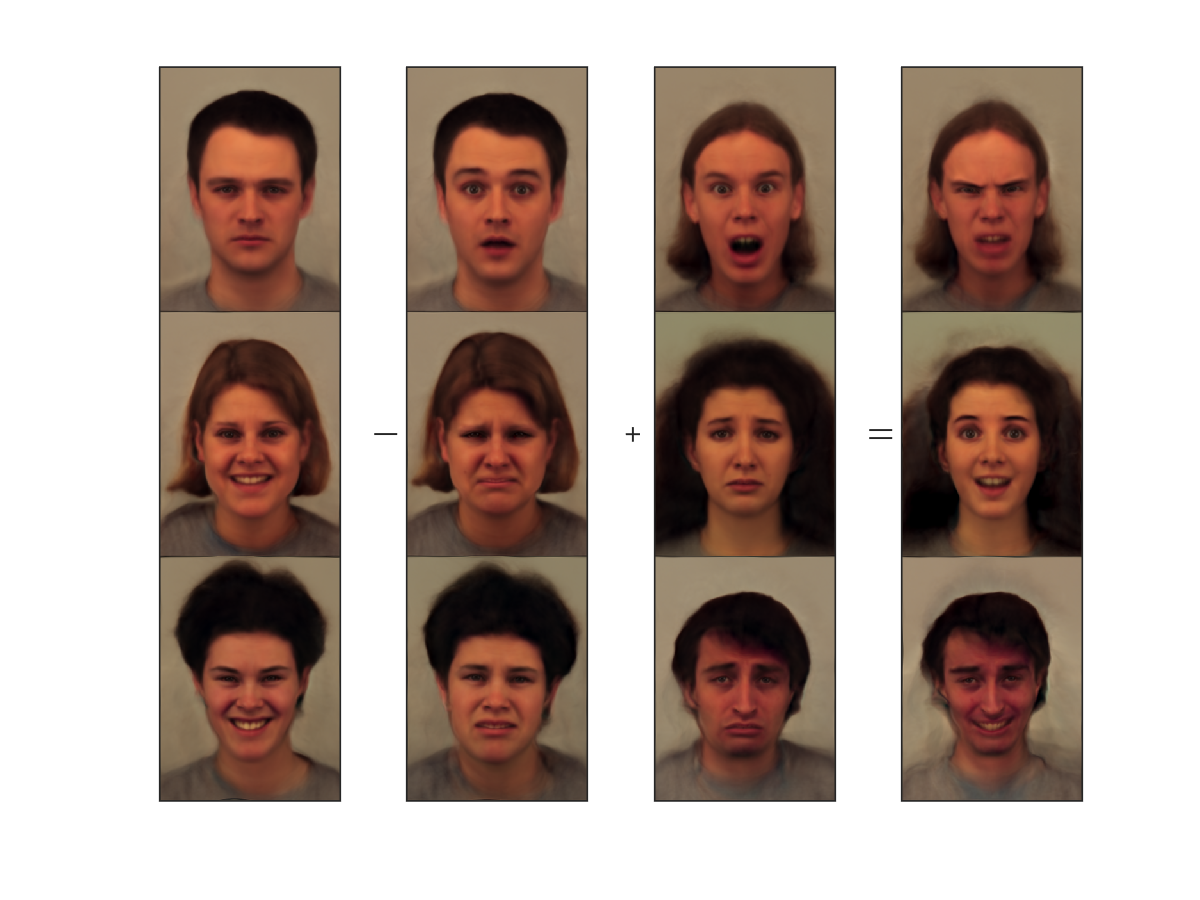,width=\textwidth}
\end{center}
\caption{
An example of simple linear additions and subtractions applied to the latent variables.
The first three columns show the full shape and appearance model fits to various faces.
Images in the right hand column were generated by making linear combinations of the latent variables that encode the images in the first three columns, and then reconstructing from these linear combinations.
\label{Fig:FaceMaths}}
\end{figure}

In many machine learning applications, it is useful to be able to model certain non-linearities in the data in a linear way, allowing more interpretable linear methods to be used while still achieving a good fit.
Following \cite{radford2015unsupervised}, this section shows that simple arithmetic on the latent variables can give intuitive results.
The first three columns of Fig. \ref{Fig:FaceMaths} show the full shape and appearance model fits to various faces.
Images in the right hand column of Fig. \ref{Fig:FaceMaths} were generated by making linear combinations of the latent variables that encode the images in the first three columns, and then reconstructing from these.
Unlike arithmetic computed in pixel space (not shown), performing arithmetic on the vectors encoding the images gives reasonably plausible results.
For medical imaging applications, it may be useful (for example) to transport disease-related changes computed from one patient, or set of patients, on to the image of a new individual \citep{modat2014simulating}.
Even when not explicitly transporting such information, it is useful to have models where the encoding can be treated in an approximately linear way.

\subsection{2D experiments with MNIST}
In this section, the behaviour of the approach using ``big data'' is assessed, which gives more of an idea of how this type of method may behave with some of the very large image datasets currently being collected.
Instead of testing on a large collection of medical images, the approach was applied to a large set of tiny images of hand-written digits.
MNIST\footnote{\url{http://yann.lecun.com/exdb/mnist/}.} \citep{lecun1998gradient} is a modified version of the handwritten digits from the National Institute of Standards and Technology (NIST) Special Database 19.
The dataset consists of a training set of 60,000 $28 \times 28$ pixel images of the digits 0 to 9, along with a testing set of 10,000 digits.
MNIST has been widely used for assessing the accuracy of machine learning approaches, and is used here as it allows behaviour of the current approach to be compared against the state-of-the-art pattern recognition methods.

In recent years, the medical imaging community has seen many of the established ``old-school'' approaches replaced by deep learning, but in doing so, ``have we thrown the baby out with the bath water?''\footnote{This was said by the late David MacKay \citep{mackay2003information} in relation to the success of kernel methods, such as support-vector machines or Gaussian processes, which, at the time, were replacing neural networks in practical applications.}.
There may still be widely used concepts from orthodox medical imaging (i.e., not deep learning) that are still useful.
In particular, geometric transformations of images are now finding their way into various machine learning approaches (e.g. \citep{hinton2011transforming,taigman2014deepface,jaderberg2015spatial}).
Much of the early work on deep learning was performed using MNIST.
Although good accuracies were achieved, the computer vision community did not take such work seriously because the images were so small.
This, however, was the early days of deep learning (i.e., before 2012), and was a sign of things to come.
This section describes an attempt to begin to reclaim some of the territory lost to deep learning.

Unlike most conventional pattern recognition approaches, the strategy adopted here is generative.
Training involves learning independent models of the ten different digits in the training set, while testing involves fitting each model in turn to each image in the test set, and performing model comparison to assess which of the ten models better explains the data.
The training stage involved learning $\hat{\boldsymbol\mu}$, $\hat{\bf W}^a$, $\hat{\bf W}^v$ and $\hat{\bf A}$ for each digit class.
A similar strategy was previously adopted by \cite{revow1996using}.
From a probabilistic perspective, the probability of the $k$th label given an image (${\bf f}$) is
\begin{align}
P(\mathcal{M}_k|{\bf f}) {}={} \frac{P({\bf f},\mathcal{M}_k)}{P({\bf f})}
{}={} \frac{\int_{\bf z} P({\bf f}|{\bf z},\mathcal{M}_k) p({\bf z}|\mathcal{M}_k) d{\bf z} P(\mathcal{M}_k)}{\sum_{l=0}^9 \int_{\bf z} P({\bf f}|{\bf z},\mathcal{M}_l) p({\bf z}|\mathcal{M}_l) d{\bf z} P(\mathcal{M}_l)}
\end{align}

The above integrals are intractable, so are approximated.
This was done by a ``Laplace approximation''\footnote{While many readers will be familiar with optimising the residual squared error (RSE), they may not understand how this differs from the evidence lower bound (ELBO). For a textbook explanation of Bayesian approaches, including the Laplace approximation, see \cite{mackay2003information}, \cite{bishop2006pattern} or \cite{murphy2012machine}.} whereby the approximate distribution of ${\bf z}$ is given by
\begin{align}
q({\bf z}) {}={} \mathcal{N}({\bf z}|\hat{\bf z},{\bf S}^{-1})
\end{align}
From this approximation, we can compute
\begin{align}
\int_{\bf z} P({\bf f}, {\bf z} | \mathcal{M}) d{\bf z} \simeq & P({\bf f}, \hat{\bf z} | \mathcal{M}) \int_{\bf z} \exp\left(-\tfrac{1}{2}
({\bf z}-\hat{\bf z})^T {\bf S} ({\bf z}-\hat{\bf z}) \right) d{\bf z}\cr
{}={} & P({\bf f}, \hat{\bf z} | \mathcal{M}) \, |{\bf S}/(2\pi)|^{1/2}
\end{align}

For each image (${\bf f}$), the mode ($\hat{\bf z}$) of $p({\bf f},{\bf z} | \mathcal{M}_k)$ was computed (see Section \ref{Sec:LatentUpdate}) by
\begin{align}
\hat{\bf z} {}={} \argmin_{{\bf z}}\left(
J({\bf f}, {\bf z}, \boldsymbol\mu, \hat{\bf W}^a, \hat{\bf W}^v)
+ \tfrac{1}{2}{\bf z}^T \left(\lambda_1 \hat{\bf A} + \lambda_2 (\hat{\bf W}^a)^T {\bf L}^a \hat{\bf W}^a + \lambda_2 (\hat{\bf W}^v)^T {\bf L}^v \hat{\bf W}^v\right){\bf z}
\right).
\end{align}
The Hessian of the objective function around this mode (Section \ref{Sec:LatentUpdate}) was used to approximate the uncertainty (${\bf S}^{-1}$).

\begin{figure}
\begin{center}
A selection of original MNIST images.\\
\epsfig{file=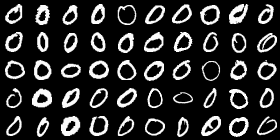,width=.19\textwidth}
\epsfig{file=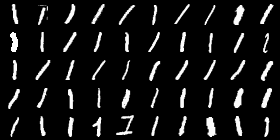,width=.19\textwidth}
\epsfig{file=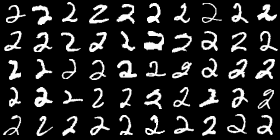,width=.19\textwidth}
\epsfig{file=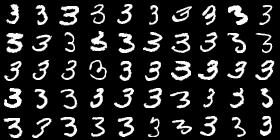,width=.19\textwidth}
\epsfig{file=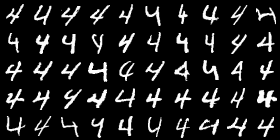,width=.19\textwidth}

\epsfig{file=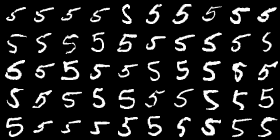,width=.19\textwidth}
\epsfig{file=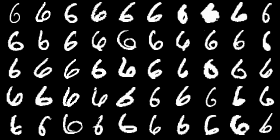,width=.19\textwidth}
\epsfig{file=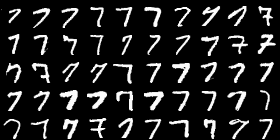,width=.19\textwidth}
\epsfig{file=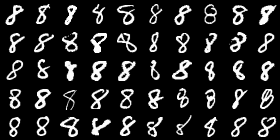,width=.19\textwidth}
\epsfig{file=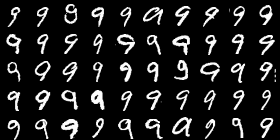,width=.19\textwidth}

Full model fit to MNIST images.\\
\epsfig{file=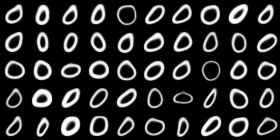,width=.19\textwidth}
\epsfig{file=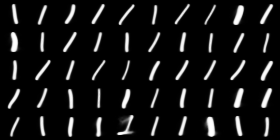,width=.19\textwidth}
\epsfig{file=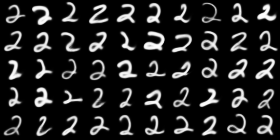,width=.19\textwidth}
\epsfig{file=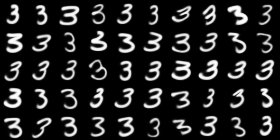,width=.19\textwidth}
\epsfig{file=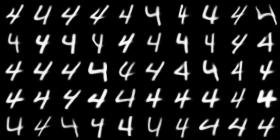,width=.19\textwidth}

\epsfig{file=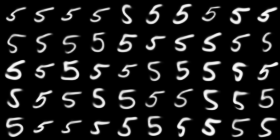,width=.19\textwidth}
\epsfig{file=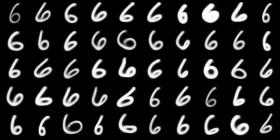,width=.19\textwidth}
\epsfig{file=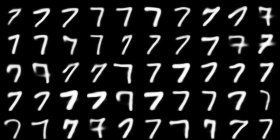,width=.19\textwidth}
\epsfig{file=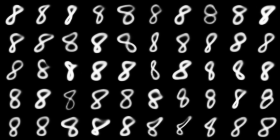,width=.19\textwidth}
\epsfig{file=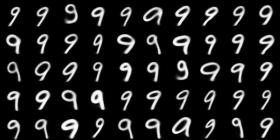,width=.19\textwidth}
\end{center}
\caption{
A random selection of digits from the first 10,000 MNIST training images, along with the model fit. In general, good alignment is achieved.\label{Fig:MNIST}}
\end{figure}

Training was done with different sized subsets (300, 500, 1,000, 3,000, 5,000, 10,000, and all 60,000) of the MNIST training data, whereas testing was always done using the 10,000 test images.
In each of the training subsets, the first of the images were always used, which generally leads to slightly different sized training sets for each of the digits.
Example images, along with the fit from the models trained using the first 10,000 images, are shown in Fig. \ref{Fig:MNIST}.
Model fitting was run for 20 iterations, using a Bernoulli likelihood with $K=16$, $\nu_0 = 16$, $\lambda = [0.95 \ 0.05]$, $\omega^a = [0.002 \ 0.2 \ 0]$, $\omega^{\mu} = N [10^{-7} \ 10^{-5} \ 0]$ and $\omega^v = [0.002 \ 0.02 \ 2 \ 0.2 \ 0.2]$.

When applied to medical images, machine learning can suffer from the curse of dimensionality.
The number of pixels or voxels in each image ($M$) is often much greater than the number of labelled images ($N$) available for training.
For MNIST, there are 60,000 training images, each containing 784 pixels, giving $N/M \simeq 75$.
In contrast, even after down-sampling to a lower resolution, a 3D MRI scan contains in the order of 20,000,000 voxels.
Achieving a similar $N/M$ as for MNIST would require about 1.5 billion labelled images, which clearly is not feasible.
For this reason, this section focuses on classification methods trained using smaller subsets of the MNIST training data.
Accuracies are compared against those reported by \cite{lee2015deeply} for their Deeply Supervised Nets, which is a deep learning approach that performs close to state-of-the-art (for 2015), particularly for smaller training sets.
Invariant scattering convolutional networks are also known to work well for smaller training sets, so some accuracies taken from \cite{bruna2013invariant} are also included in the comparison.
We are not aware of more recent papers that assess the accuracy of deep learning using smaller training sets.

\begin{figure}
\begin{center}
\epsfig{file=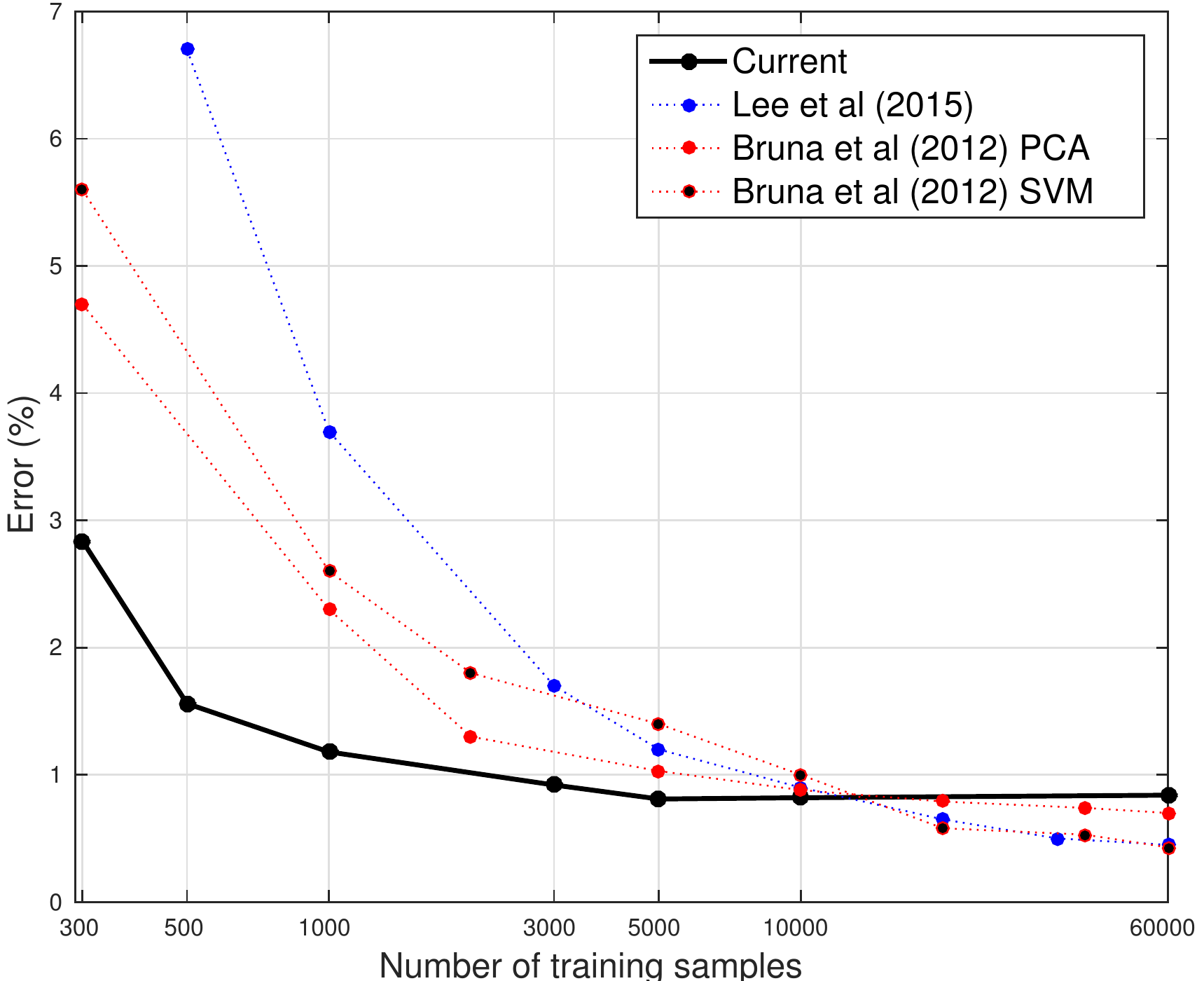,height=.45\textwidth}
\hspace{0.5cm}
\epsfig{file=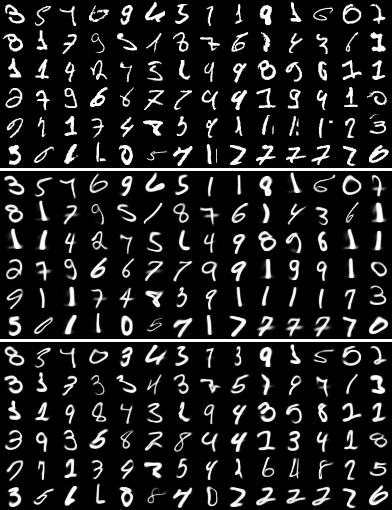,height=.45\textwidth}
\end{center}
\caption{
Left: Test errors from training the method using different sized subsets of the MNIST data (the error rate from random guessing would be 90\%).
Right: All the MNIST digits the method failed to correctly identify (after training with the full 60,000) are shown above.  These are followed by the model fits for the true digit, and then the model fits for the incorrect guess (i.e., the one with the most model evidence).
\label{Fig:MNIST_Xval}}
\end{figure}

Plots of error rate against training set size are shown in Fig. \ref{Fig:MNIST_Xval}, along with the approximate error rates from \cite{lee2015deeply} and \cite{bruna2013invariant}.
The plot shows the proposed method to be more accurate than deep learning for smaller training sets, but it is less accurate when using the full training set, as the error rate plateaus to a value of about 0.85\% for training set sizes of around 5,000 onward.
Visual assessment of the fits to the misclassified digits (Fig. \ref{Fig:MNIST_Xval}) suggests that relatively few of the failures can be attributed to registration errors.

\begin{figure}
\begin{center}
\epsfig{file=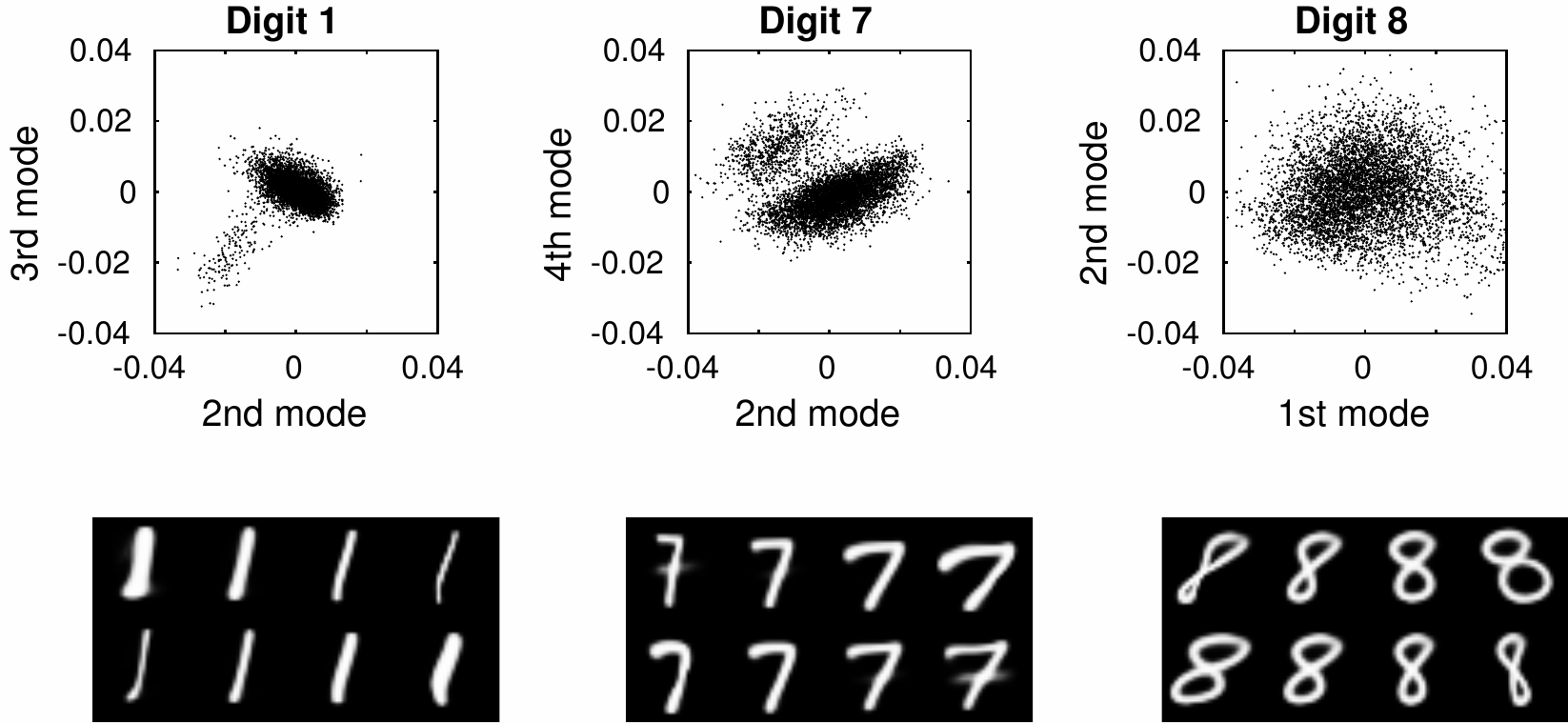,width=0.8\textwidth}
\end{center}
\caption{Illustration of the non-Gaussian distributions of the latent variables for some of the MNIST digits.  Plots of selected latent variables are shown above, with the corresponding modes of variation shown below.  Gaussian mixture models are likely to provide better models of variability than the current assumption of a single Gaussian distribution. \label{Fig:NeedGMM}}
\end{figure}

These experiments with MNIST suggest that one avenue of further work could be to elaborate on the simple multivariate Gaussian model for the distribution of latent variables.
Although accuracies were relatively good for smaller training sets, the Gaussian assumptions meant that increasing the amount of training data beyond about 5,000 examples did not bring any additional accuracy.
One example of where the Gaussian distribution fails is when attempting to deal with sevens written either with or without a bar through them, which clearly requires some form of bimodal distribution to describe (see Fig. \ref{Fig:NeedGMM}).
One approach to achieving a more flexible model of the latent variable probability density would to use a Gaussian Mixture Model (GMM) \citep{cootes1999mixture}.
This type of approach has been used for appearance models \citep{van2010capturing}.
In principle, a variational Bayesian GMM \citep{bishop2006pattern} could be used to automatically select the optimal model complexity, leading to an approach that self-tunes its complexity according to the amount and quality of of data available.
This type of model selection has previously been used for shape modelling \citep{gooya2015joint,le2017spectral}, as well as for other aspects of medical image computing.
With a more flexible model, it may be possible to achieve accuracies similar to those achieved by deep learning, but with fewer labelled training examples.
One of the aims of the Medical Informatics Platform of the HBP was to cluster patients into different sub-groups.
In addition to achieving greater accuracy (probably), incorporating a GMM over the latent variables could also lead to this clustering goal being achieved.

\subsection{Experiments with segmented MRI}
Experiments were performed using 1,913 T1-weighted MR images from the following datasets.
\begin{itemize}
\item
The \emph{IXI} dataset, which is available  under the Creative Commons CC BY-SA 3.0 license from \url{http://brain-development.org/ixi-dataset/}.
Information about scanner parameters and subject demographics are also available from the web site.  Scans were collected on three different scanners using a variety of MR sequences.
This work used only the 581 T1-weighted scans.
\item
The \emph{OASIS Longitudinal} dataset is described in \cite{marcus2010open}.  The dataset contains longitudinal T1-weighted MRI scans of elderly subjects, some of whom had dementia.  Only data from the first 82 subjects of this dataset were downloaded from \url{http://www.oasis-brains.org/}, and averages of the scans acquired at the first time point were used. 
\item
The \emph{COBRE} (Centre for Biomedical Research Excellence) dataset are available for download from \url{http://fcon_1000.projects.nitrc.org/indi/retro/cobre.html} under the Creative Commons CC BY-NC license. The dataset includes fMRI and T1-weighted scans of 72 patients with Schizophrenia and 74 healthy controls. Only the T1-weighted scans were used.  Information about scanner parameters and subject demographics is available from the web site.
\item
The \emph{ABIDE I} (Autism Brain Imaging Date Exchange) dataset was downloaded via \url{http://fcon_1000.projects.nitrc.org/indi/abide/abide_I.html} and is available under the Creative Commons CC BY-NC-SA license. There were scans from 1,102 subjects, where 531 were individuals on the Autism Spectrum. Subjects were drawn from a wide age range and were scanned at 17 different sites around the world.  All the T1-weighted scans were used, and these had a very wide range of image properties, resolutions and fields of view.  For example, many of the scans did not cover the cerebellum.
\end{itemize}

The images were segmented using the algorithm in SPM12, which uses the approach described in \cite{ashburner2005unified}, but with some additional modifications that are described in the appendices of \cite{weiskopf2011unified,malone2015accurate}.
Binary maps of grey and white matter were approximately aligned into ICBM152 space using a rigid-body transform obtained from a weighted Procrustes analysis \citep{gower1975generalized} of the deformations estimated by the segmentation algorithm.
These approximately aligned images have an isotropic resolution of 2 mm.

\subsubsection{2D experiments with segmented MRI}
It is generally easier to visualise how an algorithm is working when it is run in 2D, rather than 3D.
The examples here will be used to illustrate the behaviour of the algorithm under topological changes, when variability can not be modelled only via diffeomorphic deformations.

A single slice was extracted from the grey and white matter images of each of the 1,913 subjects, and the joint shape and appearance model was fit to the data using the settings for categorical image data.
This assumed that each voxel was a categorical variable indicating one of three tissue classes (grey and white matter, as well as background).  Each 2D image was encoded by 100 latent variables (i.e. $K=100$).
Eight iterations of the algorithm were used, with $\lambda = [0.9 \ 0.1]$, $\omega^a = [0.1 \ 16 \ 128]$, $\omega^{\mu} = N [0.0001 \ 0.01 \ 0.1]$, $\omega^v = [0.001 \ 0 \ 32 \ 0.25 \ 0.5]$ and $\nu_0 = 100$.

\begin{figure}
\begin{center}
Original data.\\
\epsfig{file=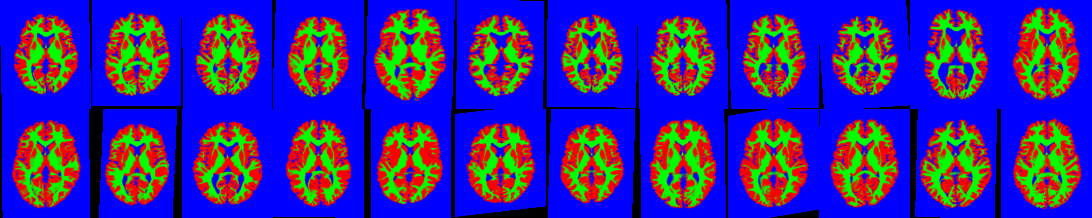,width=\textwidth}

Shape and appearance model fit.\\ 
\epsfig{file=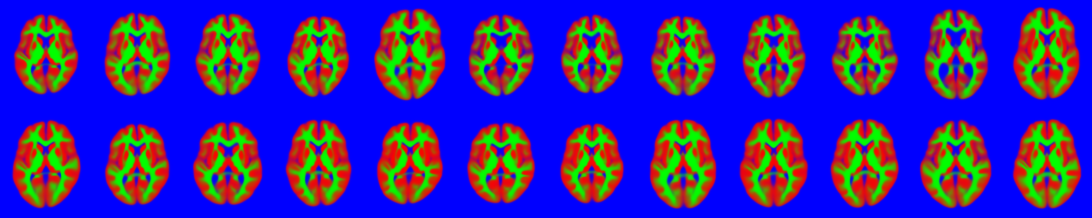,width=\textwidth}
\end{center}
\caption{
A random selection of the 2D brain image data, showing grey matter (red), white matter (green) and other (blue). Black regions indicate missing data.  Below these is the model fit to the images.\label{Fig:NIFTI2D}}
\end{figure}

\begin{figure}
\begin{center}
\epsfig{file=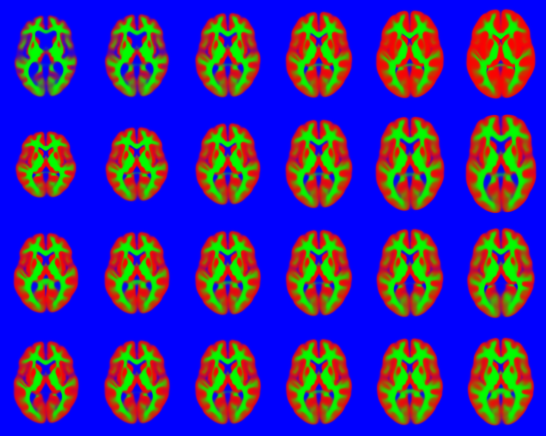,width=.49\textwidth}
\epsfig{file=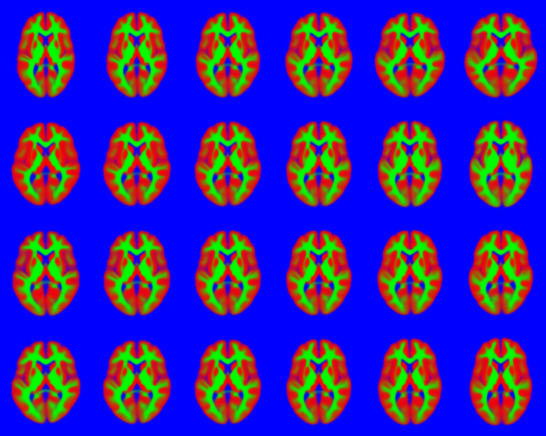,width=.49\textwidth}
\end{center}
\caption{
First eight (out of a total of 100) modes of variability found from the 2D brain image dataset, shown at -5, -3, -1, +1, +3 \& +5 standard deviations. Note that these modes encode some topological changes, in addition to changes in shape.\label{Fig:NIFTI2Dmodes}}
\end{figure}

Some model fits are shown in Fig. \ref{Fig:NIFTI2D}, and the principal modes of variability are shown in Fig. \ref{Fig:NIFTI2Dmodes}.
As can be seen, these images are reasonably well modelled, although achieving a similar quality of fit for the full 3D data would probably require about 1,000 ($100^{3/2}$) variables.
Fitting a model of this size would require many more than the 1,900 subjects in this dataset.
Note that the topology of the images may differ, which (by definition\footnote{Topology is concerned with properties that are preserved following diffeomorphic deformations (see \url{https://en.wikipedia.org/wiki/Topology}).}) is not something that can be modelled by diffeomorphisms alone.
The inclusion of the appearance model allows these topology differences to be more accurately captured.

\begin{figure}
\begin{center}
\epsfig{file=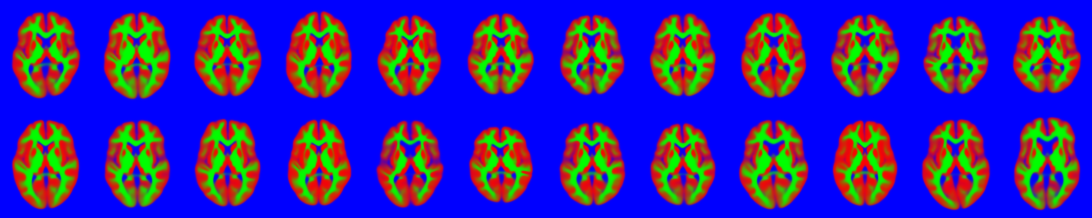,width=\textwidth}
\epsfig{file=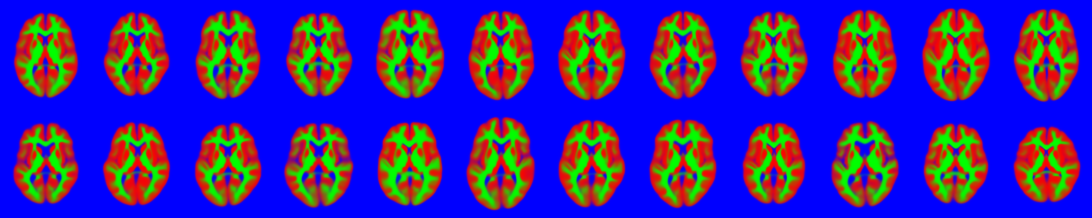,width=\textwidth}
\end{center}
\caption{
Randomly generated slice through brain images. These images were constructed by using randomly assigned latent variables.  Note that the top set of images uses the same random variables as the bottom set, except they are multiplied by $-1$. This means that one set is a sort of ``opposite'' of the other. For example, if a brain in the upper set has large ventricles, then the corresponding brain in the lower set will have small ventricles.\label{Fig:NIFTI2Drandom}}
\end{figure}

\subsubsection{Imputing missing data \label{Sec:Missing}}
The ability to elegantly handle missing data is a useful requirement for mining hospital scans.
These often have limited fields of view, and may miss out parts of the brain that are present in other images.
The objective here is to demonstrate that a reasonable image factorisation can be learned, even when some images in the dataset may not have full organ coverage.

\begin{figure}
\begin{center}
Original data.\\
\epsfig{file=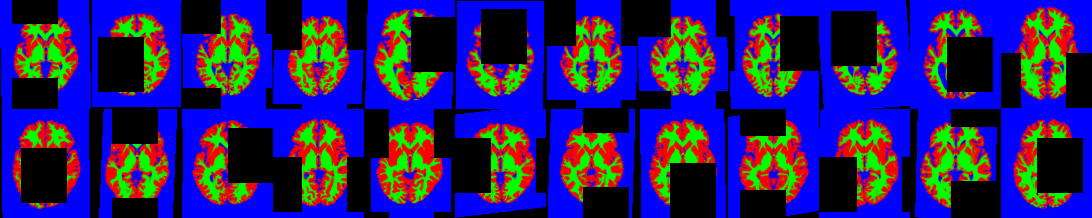,width=\textwidth}

Missing data filled in.\\ 
\epsfig{file=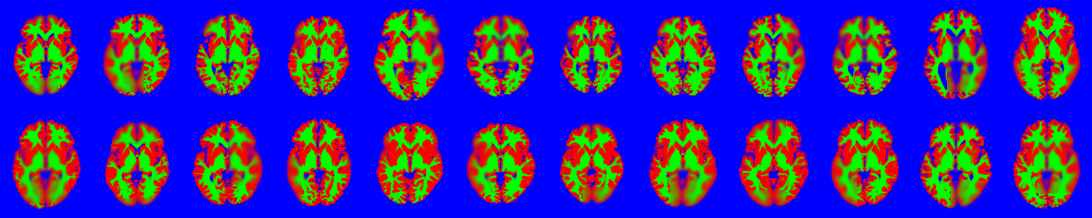,width=\textwidth}
\end{center}
\caption{
A random selection of the 2D brain image data showing the location of missing data.  The attempt to fill in the missing information is shown below.  These may be compared against the original images shown in Fig. \ref{Fig:NIFTI2D}. \label{Fig:NIFTI2Dmissing}}
\end{figure}

This experiment used the same slice through the data as above, and a rectangle covering 25\% of the area of the images was placed randomly in each and every image of the training set (wrapping around at the edge of the field of view), and the intensities within these rectangles set to NaN (``not a number'' in the IEEE 754 floating-point standard).
The algorithm was trained, using the same settings as described previously, on the these modified images.
Although imputed missing values may not be explicitly required, they do provide a useful illustration of how well the model works in less than ideal situations.
Fig. \ref{Fig:NIFTI2Dmissing} shows a selection of the images with regions set to NaN, and the same images with the missing values predicted by the algorithm.

The ability to handle missing data allows cross-validation to be used to determine the accuracy of a model, and how well it generalises.
In addition to the joint shape and appearance model, this work also allows simplified versions to be fitted that involve only shape (i.e., not using ${\bf W}^a$, as in \cite{zhang2015bayesian}) or in a form that varies only the appearance (i.e. not using ${\bf W}^v$).
In addition, this work also includes a version where different sets of latent variables control the shape and appearance.
Here, there were 30 variables to control appearance, and 70 to control shape.
The aim was to compare the four models by assessing how well they are able to predict data that was unavailable to the model during fitting.
This gives us ground truth with which to compare the models' predictions, and is essentially a form of cross-validation procedure.
Accuracy was measured by the log-likelihood of the ground truth data, which was computed only for pixels that the models did not have access to during training.

\begin{figure}
\begin{center}
\epsfig{file=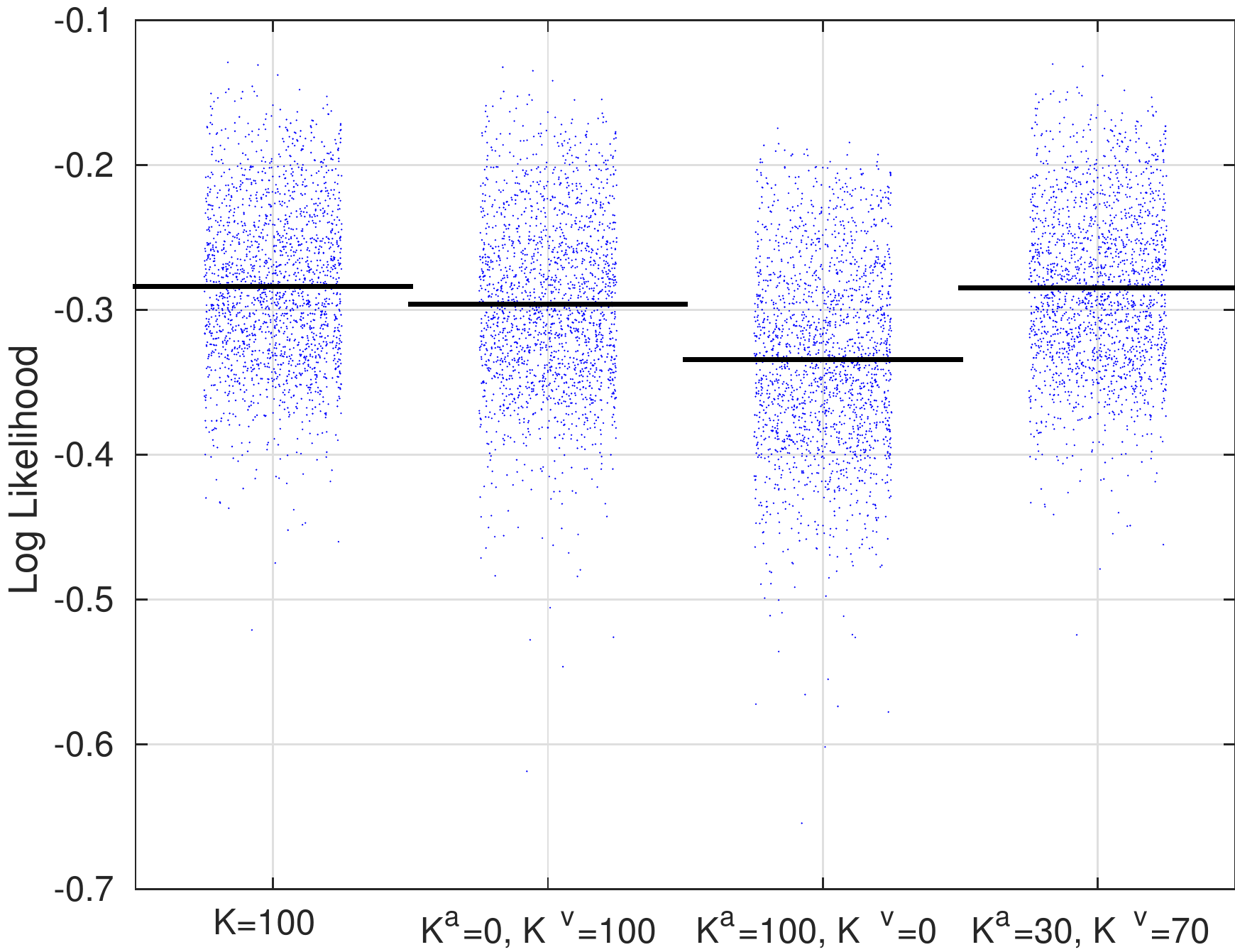,width=.45\textwidth}\hspace{1cm}
\epsfig{file=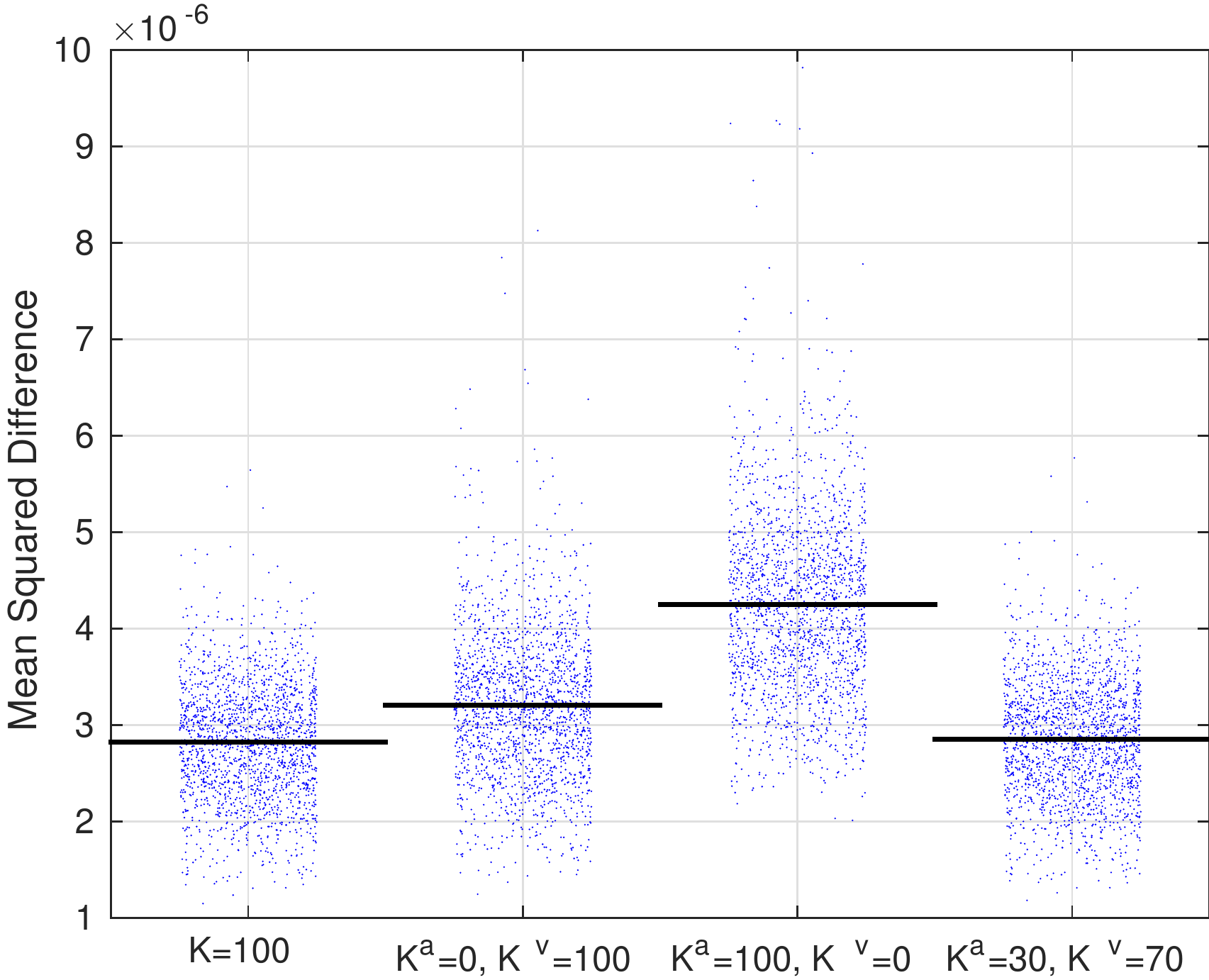,width=.45\textwidth}
\end{center}
\caption{Cross-validation accuracy measures based on predicting the left-out patches of the images. The blue dots show the mean value for each of the 1,913 images, whereas the horizontal bars show the mean values overall.\label{Fig:missingXval}}
\end{figure}

The results of the cross-validation are shown in Fig. \ref{Fig:missingXval}, and clearly show that the models that combine both shape and appearance have greater predictive validity than either the shape or appearance models alone.
Mean squared errors from the different models are also presented, and these exhibit the same general pattern.
Although the difference was small, the best results were from the model where each latent variable controls both shape and appearance, rather than when they are controlled separately.
Both of these combined models outperformed models of only shape or only appearance variability.
Note that changes to hyper-parameter settings, etc. may improve accuracies further.

\subsubsection{3D experiments with segmented MRI}
The aim of this section was to apply the method to a large set of 3D images, and use the resulting latent variables as features for pattern recognition.
For this, a version of the model was used whereby some latent variables controlled appearance, whereas others controlled shape.
The motivation for this was that it allows the different types of features to be differentially weighted when they are used to make predictions.

\begin{figure}
\begin{center}
Appearance model part.\\
\epsfig{file=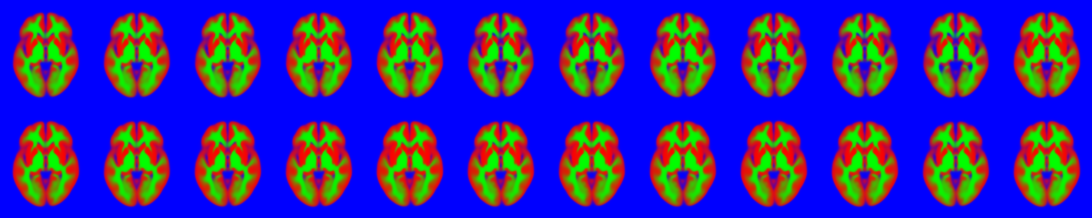,width=\textwidth}

Shape model part.\\ 
\epsfig{file=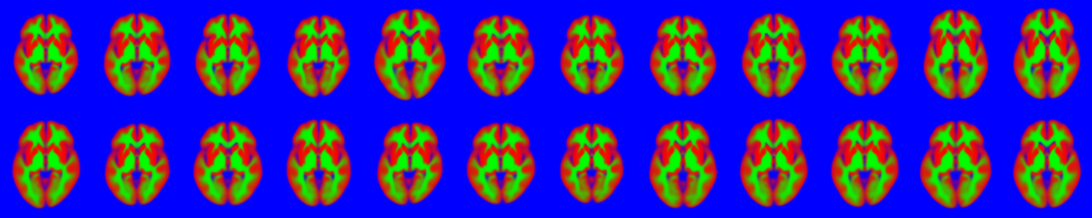,width=\textwidth}

Both shape and appearance.\\ 
\epsfig{file=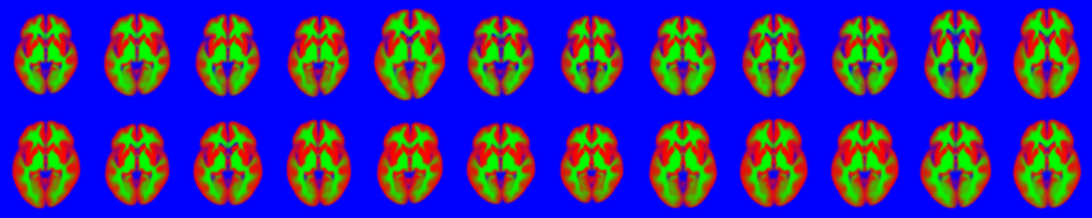,width=\textwidth}

\end{center}
\caption{
An illustration of slice 40 (one out of 91 slices) of the model fits to the random selection of 3D images.  These may be compared against the original images shown in Fig. \ref{Fig:NIFTI2D}.
\label{Fig:NIFTI3D}}
\end{figure}

The algorithm was run on the full 3D dataset, using 70 variables to control shape and 30 to control appearance.  Eight iterations were used, with $\lambda = [1 \ 1]$, $\omega^a = [0.01 \ 1 \ 50]$, $\omega^{\mu} = N [0.00001 \ 0.01 \ 0.1]$ and $\omega^v = [0.001 \ 0 \ 10 \ 0.1 \ 0.2]$.
The resulting model fits are shown in Fig. \ref{Fig:NIFTI3D}, as well as reconstructions using only the appearance part of the model or the shape model part.
This result can be compared against the 2D model fit shown in Fig. \ref{Fig:NIFTI2D}.
There are two reasons why the 3D fit explains a smaller proportion of the variability than for the 2D examples. The first is that the 3D model fit uses different variables to control shape and appearance, meaning that each can explain slightly less of the variability.
The second reason is simply that it is a 3D fit, so that there is a great deal more variability to explain.

The main objective of this work is to extract features from sets of medical images, which are effective for machine learning applications.
Here, a five-fold cross-validation is used to assess the effectiveness of these features.
Machine learning used a linear Gaussian process classification procedure, which is essentially equivalent to a Bayesian approach to logistic regression.
The implementation was based on the method for binary classification using expectation propagation described in \cite{rasmussen2006gaussian}.
For the COBRE dataset, classification involved separating controls from patients with schizophrenia.
Similarly, the analysis of the ABIDE dataset involved identifying those subjects on the autism spectrum, with features orthogonalised with respect to the different sites.
Classification involved three hyper-parameters, which weighted the contributions from shape features, appearance features and a constant offset.

\begin{figure}
\begin{center}
\epsfig{file=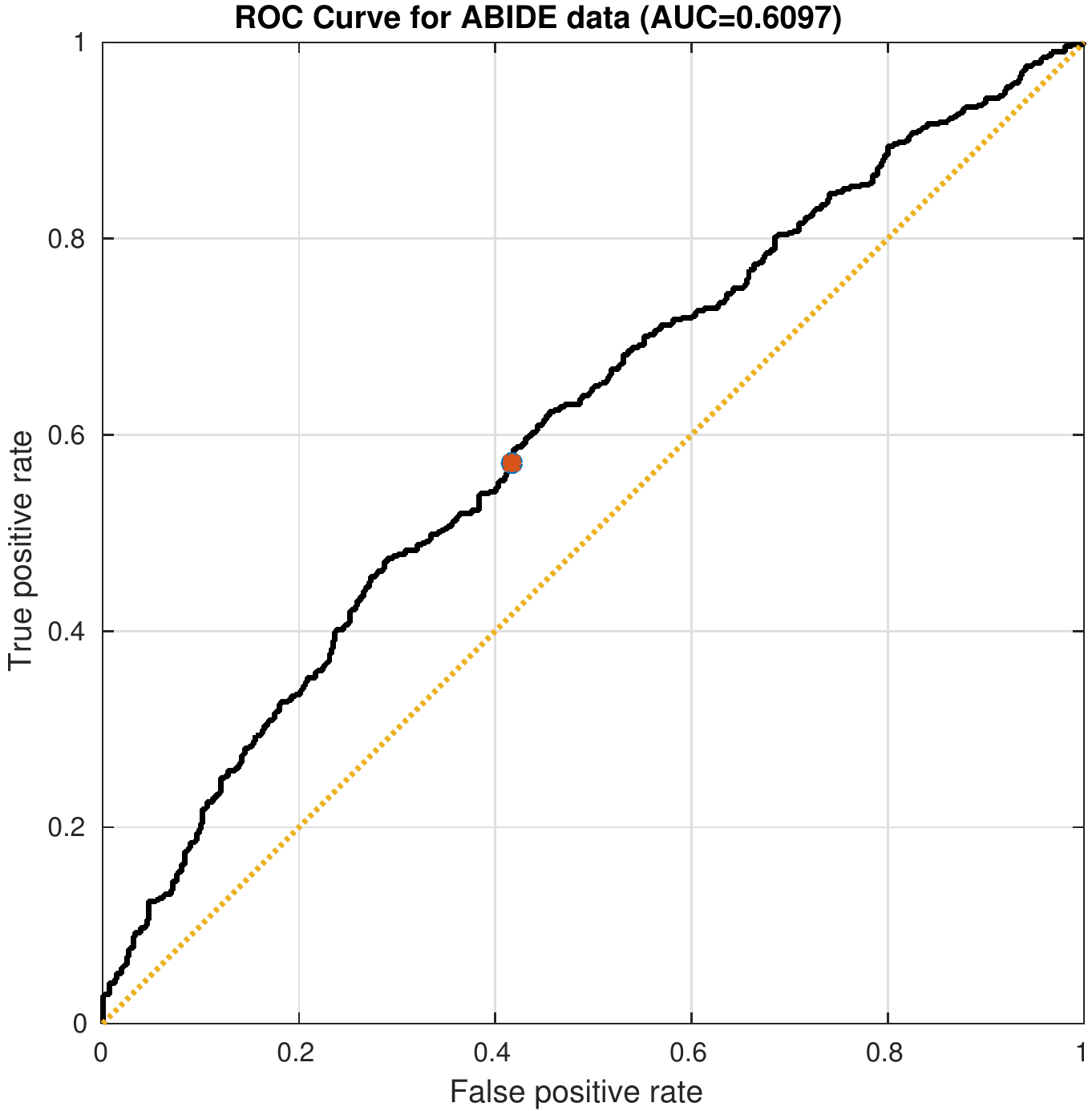,width=.45\textwidth}\hspace{1cm}
\epsfig{file=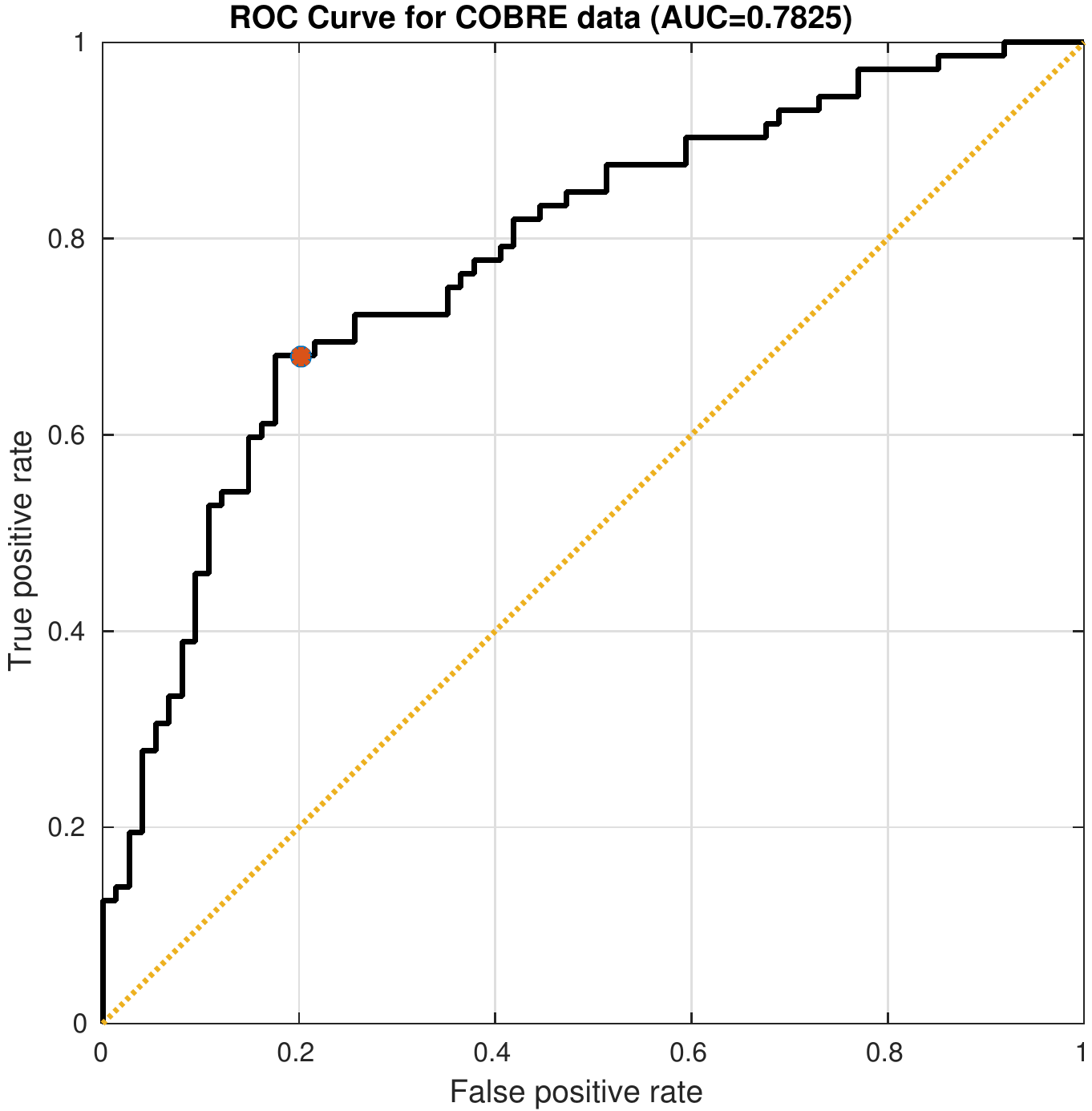,width=.45\textwidth}
\end{center}
\caption{ROC curves from five-fold cross-validation accuracies from the ABIDE and COBRE data. Red dots show the point on the curve where the classification gives probabilities of 0.5.\label{Fig:ROC}}
\end{figure}

ROC curves are shown in Fig. \ref{Fig:ROC}.
For ABIDE, the accuracy was 57.6\%.  While this is not especially high, it is close to the accuracy reported by others who have applied machine learning to the T1-weighted scans.
For example, \cite{ghiassian2016using} reported 60.1\% classification accuracy with the same data using a support-vector machine with radial basis functions.
The accuracy achieved for the COBRE dataset was 74.7\%, which is similar to the  69.7\% accuracy reported by \cite{cabral2016classifying} using COBRE.  \cite{nieuwenhuis2012classification} achieved 71.4\% accuracy for separating controls from subjects with schizophrenia, but using a different (larger) dataset of T1-weighted scans.
Both results were comparable to those obtained by \cite{monte2018comparison}.

\section{Discussion}
This work presents a very general generative framework that may have widespread use within the medical imaging community, particularly for those situations where conventional image registration approaches are more likely to fail.
As in many image analysis procedures, there is a certain amount of circularity arising from the dependencies among variables.
These circularities are not handled well by using a single pass through a conventional pipeline.
In contrast, by formulating the problem within a generative modelling framework, the algorithm is better able to deal with them.
Because of its generality, the model we presented should provide a good starting point for a number of avenues of further development, in addition to those previously mentioned that relate to going beyond a simple Gaussian distribution for the latent variables.

Most image analysis applications have a number of settings to be tuned, and the current approach is no exception.
Although this tuning is rarely discussed in papers, the settings can have quite a large impact on any results.
The effects of different forms of regularisation used for registration are illustrated in \cite{ashburner2012symmetric}.
The choice of settings plays a more important role when training using only a small number of images.
With more images, the contribution of these priors becomes less influential.
One aspect of the work that may need additional attention is the setting of $\lambda_1$ and $\lambda_2$.
The ideal settings would be $\lambda_1=1$ and $\lambda_2=0$, but in practice, greater regularisation is required in order to achieve good results.
A plausible explanation for this would be that assumptions of i.i.d. noise are not generally met, so a ``virtual decimation factor'', which accounts for correlations among residuals, may need to be accounted for \citep{groves2011linked}.
The fact that the approach is not fully Bayesian (i.e., it only makes point estimates of many parameters and latent variables, rather than properly accounting for their uncertainty) is likely to be another reason why additional regularisation is needed.
Greater regularisation also seems to help avoid local optima, suggesting that further investigation into coarse-to-fine strategies may be warranted.

For computational reasons, the uncertainty with which the latent variables (${\bf z}$) are estimated is not treated in a properly Bayesian way.
In theory, more accurate results would be achieved by marginalising with respect to these latent variables.
This strategy is relatively straightforward for PCA with simple Gaussian noise assumptions \citep{tipping1999probabilistic_pri}, but becomes more difficult for other types of data.
Monte-Carlo approaches \citep{allassonniere2010construction,allassonniere2015bayesian,zhang2013probabilistic} are currently too slow for routine use on large datasets, although a reasonable alternative might be to use variational Bayesian methods.
For example, \cite{tipping1999probabilistic_vis} used a local variational approximation of the log partition function to do a type of generalised PCA of binary data.
\cite{elhabian2017label} used a related approach for modelling the shape variability of binary images, although this work used a different local approximation.
Local variational approaches appear to be more difficult to extend to categorical data with more than two classes, and the image deformation part of the model would probably lead to many further difficulties for variational Bayesian methods.

There are a number of directions in which the current work could be extended.
One of the first areas we plan to investigate is to integrate the approach with current strategies for generating tissue probability maps \citep{bhatia2007groupwise,purwani2014ensemble,blaiotta2018generative}.
Rather than first segmenting images into different tissue classes and then fitting the proposed model to these, it would be more elegant to properly combine the approaches into a single unified generative model of the original images.
In principle, this strategy could lead to more robust segmentation because the ways in which tissue priors are allowed to vary are constrained to be more biologically plausible.

Another avenue is to allow some shape variability beyond what can be encoded by the first few eigenmodes.
For example, \cite{balbastre2018diffeomorphic} combined the eigenmode representation with a model of additional shape variability, giving a framework that is conceptually related to that of \cite{allassonniere2007towards}, as this allows a covariance matrix over velocity fields to be defined and optimised.
Approaches building on \cite{sang2012full}, who proposed to learn the residual covariance by assuming finite support (and thus sparsity) of the underlying covariance function, might also be potentially interesting.

The framework would also generalise further for handling paired or multi-view data, which could add a degree of supervision to the method.
There have been a number of publications on generating age- or gender-specific templates, or on geodesic regression approaches \citep{niethammer2011geodesic,fletcher2013geodesic} for modelling trajectories of ageing.
Concepts from joint matrix factorisation approaches, such as canonical correlation analysis \citep{bach2005probabilistic,klami2013bayesian}, could be integrated into the current work, and these could be used to allow the model fitting to be informed by age, gender, disease status etc.

\subsubsection*{Acknowledgements}
This project has received funding from the European Union’s Horizon 2020 Research and Innovation Programme under Grant Agreement No. 720270 (HBP SGA1).
The Wellcome Centre for Human Neuroimaging is supported by core funding from the Wellcome Trust [grant number 203147/Z/16/Z].  Funding for the OASIS dataset came from the following grants: P50 AG05681, P01 AG03991, R01 AG021910, P20 MH071616, U24 RR021382.
Funding for the ABIDE I dataset came from a variety of sources, which include NIMH (K23MH087770 and R03MH096321), the Leon Levy Foundation, Joseph P. Healy and the Stavros Niarchos Foundation to the Child Mind Institute.


\appendix
\section{Algorithm \label{Sec:Algorithm}}
A highly simplified version of the factorisation algorithm is shown in Algorithm \ref{Alg:PG}.
It is an expectation maximisation approach, which involves alternating between computing the shape and appearance basis functions (plus a few other variables -- \emph{M-step}), and computing the expectations of the latent variables (\emph{E-step}).
For better convergence of the basis function updates of the M-step, an orthogonalisation step is included in the algorithm.

The M-step relies on Gauss-Newton updates of three elements: the mean template ($\boldsymbol\mu$), shape subspace (${\bf W}^a$) and appearance subspace (${\bf W}^v$).
These updates have the general form of ${\bf w} \gets {\bf w} - ({\bf H} + {\bf L})^{-1}({\bf g} + {\bf L}{\bf w})$, where ${\bf L}$ is a very sparse Toeplitz or circulant matrix encoding spatial regularisation, and ${\bf H}$ encodes a field of small matrices that are easy to invert.
The full-multigrid method, described in \cite{ashburner2007fast}, is particularly well suited to solving this type of problem.

The E-step involves updating the distributions of the latent variables (${\bf Z}$) and Gaussian prior (${\bf A}$).
To break the initial symmetry, the latent variables are all initialised randomly, while ensuring that $\hat{\bf Z}\hat{\bf Z}^T$ is orthonormal.
This is easier than random initialisation of $\hat{\bf W}^a$ and $\hat{\bf W}^v$, which are instead initialised to zero. 

In most of this appendix, matrices are written in bold upper-case (e.g., ${\bf W}^a$, ${\bf Z}$, etc).
In the computations, images are treated as vectors. These are written as lower-case bold, which includes the notation for individual columns of various matrices (e.g., ${\bf w}^a_k$ denotes the $k$th column of ${\bf W}^a$, ${\bf z}_n$ denotes the $n$th column of ${\bf Z}$, etc).
Scalars are written in italic, with dimensions in upper-case.
Estimates or expectations of parameters are written with a circumflex (e.g., $\hat{\bf Z}$).
Collections of vectors may be conceptualised as matrices, so are written in bold-upper-case (e.g., ${\bf G}^a$, where individual vectors are ${\bf g}^a_k$).
Collections of matrices are written in ``mathcal'' font (e.g., $\mathcal{H}^a$, where individual matrices are ${\bf H}^a_{kk}$).
There is no systematic rule about the use of Greek letters, although sometimes warping an image may be written as ${\bf a}(\psi)$ and at other times, warping may be conceptualised as matrix multiplication and written as $\boldsymbol\Psi {\bf a}$.
The matrix transpose operation is denoted by the ``$^T$'' superscript (as in $\boldsymbol\Psi^T$).
Creating a diagonal matrix from a vector (as in $\diag(\exp {\bf q})$), as well as treating the diagonal elements of a matrix as a vector (as in $\diag({\bf Q})$) are both denoted by ``$\diag$''. 
The trace of a matrix (sum of diagonal elements) is denoted by ``$\Tr$''.
Sometimes, gradients of an image are required.  In 3D, the three components of the spatial gradient of ${\bf a}$ would be denoted by $\nabla_1 {\bf a}$, $\nabla_2 {\bf a}$ and $\nabla_3 {\bf a}$.

Comments in Algorithm \ref{Alg:PG} saying ``Dist'' indicate which steps should be modified for running within a distributed privacy-preserving framework.
The idea here is that the main procedure would be run on the ``master'' computer, whereas various functions would be run on the ``worker'' machines on which the data reside.
These workers would only pass aggregate data back to the master, whereas the latent variables, which explicitly encode information about individuals, would remain on the workers.
As the algorithm is described here, the images (${\bf F}$) and expectations of the latent variables $\hat{\bf Z}$ are passed back and forth between the master and workers, but this need not be the case.
If these data and variables were all to reside on the worker machines, the master machine would still be able to run using only the aggregate data.

For simplicity, Algorithm \ref{Alg:PG} does not include functions for computing variances ($\sigma^2$ used by the Gaussian noise model), etc., and these variables are not shown to be passed to the various functions that use them.
However, it should be easy to see how these changes would be incorporated in practice.

Also, the illustration does not show any steps requiring the objective function, which include various backtracking line-searches to ensure that parameter updates cause the objective function to improve each time.
In practice, the algorithm is run for a fixed number of iterations, although the log-likelihood could be used to determine when to stop.

\begin{algorithm}
\caption{Shape and appearance model}
\label{Alg:PG}
\begin{algorithmic}
    \State Initialize variables.
    \Repeat
        \State ${\bf g}^{\mu}, {\bf H}^{\mu} \gets \text{MeanDerivatives}({\bf F}, \hat{\bf Z}, \hat{\boldsymbol\mu}, \hat{\bf W}^a, \hat{\bf W}^v)$ \Comment Dist
        \State $\hat{\boldsymbol\mu} \gets \hat{\boldsymbol\mu} - ({\bf H}^{\mu} + {\bf L}^{\mu})^{-1} ({\bf g}^{\mu} + {\bf L}^{\mu} \hat{\boldsymbol\mu})$
        \State
        \State ${\bf G}^v, \mathcal{H}^v \gets \text{ShapeDerivatives}({\bf F}, \hat{\bf Z}, \hat{\boldsymbol\mu}, \hat{\bf W}^a, \hat{\bf W}^v)$ \Comment Dist
        \For{$k=1...K^v$}
            \State $\hat{\bf w}^v_k \gets \hat{\bf w}^v_k - ({\bf H}^v_{kk} + (\lambda_1 N + \lambda_2 c^z_{kk}){\bf L}^v)^{-1} ({\bf g}^v_k + (\lambda_1 N + \lambda_2 c^z_{kk}){\bf L}^v \hat{\bf w}^v_k)$
        \EndFor
        \State
        \State ${\bf G}^a, \mathcal{H}^a \gets \text{AppearanceDerivatives}({\bf F}, \hat{\bf Z}, \hat{\boldsymbol\mu}, \hat{\bf W}^a, \hat{\bf W}^v)$ \Comment Dist
        \For{$k=1...K^a$}
            \State $\hat{\bf w}^a_k \gets \hat{\bf w}^a_k - ({\bf H}^a_{kk} + (\lambda_1 N + \lambda_2 c^z_{kk}){\bf L}^a)^{-1} ({\bf g}^a_k + (\lambda_1 N + \lambda_2 c^z_{kk}){\bf L}^a \hat{\bf w}^a_k)$
        \EndFor
        \State
        \State ${\bf C} \gets (\hat{\bf W}^v)^T {\bf L}^v \hat{\bf W}^v + (\hat{\bf W}^a)^T {\bf L}^a \hat{\bf W}^a$
        \State $\hat{\bf Z}, {\bf S}, {\bf C}^z \gets \text{UpdateLatentVariables}({\bf F}, \hat{\bf Z}, \hat{\boldsymbol\mu}, \hat{\bf W}^a, \hat{\bf W}^v, \lambda_1 \hat{\bf A} + \lambda_2 {\bf C})$ \Comment Dist
        \State
        \State ${\bf T} \gets \text{OrthogonalisationMatrix}({\bf C}, {\bf C}^z, {\bf S}, N)$
        \State $\hat{\bf W}^a \gets \hat{\bf W}^a {\bf T}^{-1}$
        \State $\hat{\bf W}^v \gets \hat{\bf W}^v {\bf T}^{-1}$
        \State ${\bf C}^z \gets {\bf T} {\bf C}^z {\bf T}^T$
        \State ${\bf S} \gets {\bf T}{\bf S}{\bf T}^T$
        \State $\hat{\bf Z} \gets {\bf T}\hat{\bf Z}$ \Comment Dist
        \State
        \State $\hat{\bf A} \gets (N+\nu_0)({\bf C}^z + {\bf S} + \boldsymbol\Lambda_0^{-1})^{-1}$
    \Until{convergence}
\end{algorithmic}
\end{algorithm}

\subsection{Updating the mean ($\hat{\boldsymbol\mu}$)}
From Eqn. \ref{Eqn:Objective}, we see that a point estimate of the mean ($\boldsymbol\mu$) may be computed by
\begin{align}
\hat{\boldsymbol\mu} {}={} \argmin_{\boldsymbol\mu} \left(
    \tfrac{1}{2} {\boldsymbol\mu}^T {\bf L}^\mu {\boldsymbol\mu}
    + \sum_{n=1}^N J({\bf f}_n, \hat{\bf z}_n, {\boldsymbol\mu}, \hat{\bf W}^a, \hat{\bf W}^v)
\right).
\end{align}

In practice, this log probability is not fully maximised with respect to $\boldsymbol\mu$ at each iteration.
Instead, $\hat{\boldsymbol\mu}$ is updated by a single Gauss-Newton iteration.
This requires gradients and Hessians computed as shown in Algorithm \ref{Alg:mean}, which simply involves summing over those computed for the individual images.
A small amount of regularisation is used for the estimate of the mean.
One of the reasons for using this is that it alleviates some of the problems with the gradients of Jacobian-weighted averaging that was pointed out in Appendix B of \cite{ashburner2012symmetric}, thus leading to better convergence.
This is especially important in situations where it can help to smooth over some of the effects of missing data.

\begin{algorithm}
\caption{Computing gradients and Hessians for mean}
\label{Alg:mean}
\begin{algorithmic}
\Function{MeanDerivatives}{${\bf F}, \hat{\bf Z}, \hat{\boldsymbol\mu}, \hat{\bf W}^a, \hat{\bf W}^v$}
    \State ${\bf g}^{\mu} = 0$, ${\bf H}^{\mu} = 0$
    \For{$n=1...N$}
        \State ${\bf a}  \gets \hat{\boldsymbol\mu} + \hat{\bf W}^a \hat{\bf z}_n$
        \State $\boldsymbol\Psi \gets \text{Shoot}(\hat{\bf W}^v \hat{\bf z}_n)$
        \State ${\bf g}', {\bf H}' \gets \text{LikelihoodDerivatives}({\bf f}_n, {\bf a}, \boldsymbol\Psi)$
        \State ${\bf g}^{\mu} \gets {\bf g}^{\mu} + {\bf g}'$
        \State ${\bf H}^{\mu} \gets {\bf H}^{\mu} + {\bf H}'$
    \EndFor
    \State \Return ${\bf g}^{\mu}, {\bf H}^{\mu}$
\EndFunction
\end{algorithmic}
\end{algorithm}

\subsection{Likelihood derivatives}
The algorithm can be run using a number of different appearance models, and the gradients and Hessians involved in the Gauss-Newton updates depend upon the one used.

\subsubsection{Gaussian model}
Algorithm \ref{Alg:likelihood} shows derivatives for the Gaussian noise model.
For a single voxel, this is based on
\begin{align}
J_{L_2} & {}={} \tfrac{1}{2} \ln(2\pi) + \tfrac{1}{2} \ln \sigma^2 + \tfrac{1}{2\sigma^2}||f - a'||_2^2 \\
\frac{d J_{L_2}}{d a'} & {}={} \tfrac{1}{\sigma^2} (a' - f) \text{ and } \frac{d^2 J_{L_2}}{d a'^2} = \tfrac{1}{\sigma^2}
\end{align}
For voxels where data is missing, both $J_{L_2}$ and $\tfrac{d J_{L_2}}{d a'}$ are assumed to be zero.
Using matrix notation, the objective function for an image is therefore
\begin{align}
J' {}={} \tfrac{1}{2 \sigma^2} (\boldsymbol\Psi {\bf a} - {\bf f})^T(\boldsymbol\Psi {\bf a} - {\bf f}) + \tfrac{M}{2}(\ln(\sigma^2) + \ln(2\pi)).
\end{align}
The gradients and Hessians, with respect to variations in ${\bf a}$, are
\begin{align}
{\bf g}' {}={} & \boldsymbol\Psi^T \left(\tfrac{1}{\sigma^2} (\boldsymbol\Psi {\bf a} - {\bf f}) \right)\\
{\bf H}' {}={} & \tfrac{1}{\sigma^2} \boldsymbol\Psi^T \boldsymbol\Psi
\end{align}
In practice, the Hessian (${\bf H}'$) is approximated by a diagonal matrix
\begin{align}
{\bf H}' {}\simeq{} & \diag\left( \boldsymbol\Psi^T {\bf 1}\tfrac{1}{\sigma^2}\right)
\end{align}
where ${\bf 1}$ is a vector of ones.
This approximation works in the optimisation because all rows of $\boldsymbol\Psi$ sum to 1, so for any vector ${\bf d}$ of the right dimension, the rows of $\boldsymbol\Psi^T \diag({\bf d}) \boldsymbol\Psi$ sum to $\boldsymbol\Psi^T {\bf d}$.
Because (for trilinear interpolation) all elements of $\boldsymbol\Psi$ are greater than or equal to zero, so if all elements of ${\bf d}$ are non-negative, then all eigenvalues of $\diag\left( \boldsymbol\Psi^T {\bf d}\right) - \boldsymbol\Psi^T \diag({\bf d}) \boldsymbol\Psi$ are greater than or equal to zero.
Non-negative eigenvalues ensure that the approximation to the Hessian is positive semi-definite, and thus the optimisation aims for a minimum, rather than a maximum.

\begin{algorithm}
\caption{Likelihood derivatives for Gaussian noise model}
\label{Alg:likelihood}
\begin{algorithmic}
\Function{LikelihoodDerivatives}{${\bf f}, {\bf a}, \boldsymbol\Psi$}
    \State $J' \gets \tfrac{1}{2 \sigma^2} || \boldsymbol\Psi {\bf a} - {\bf f} ||^2 + \tfrac{M}{2}(\ln(\sigma^2) + \ln(2\pi)) $ \Comment If needed
    \State ${\bf g}' \gets \boldsymbol\Psi^T \left(\tfrac{1}{\sigma^2} (\boldsymbol\Psi {\bf a} - {\bf f}) \right)$
    \State ${\bf H}' \gets \diag\left( \boldsymbol\Psi^T \left(\tfrac{1}{\sigma^2}{\bf 1} \right) \right)$ \Comment where ${\bf 1}$ is an array of ones
    \State \Return $J', {\bf g}', {\bf H}'$
\EndFunction
\end{algorithmic}
\end{algorithm}

\subsubsection{Binary model}
Slight modifications are made to the algorithm in order to use the Bernoulli noise model with the sigmoidal squashing function.
For each voxel, the negative log-likelihood is
\begin{align}
J_{Bern} & {}={} -\left(f a' + \ln s(-a')\right) \text{, where } s(a') = \frac{1}{1+\exp(-a')}.
\end{align}
Modifications are made to the gradient and Hessian of Algorithm \ref{Alg:likelihood}, based on the derivatives
\begin{align}
\frac{d J_{Bern}}{d a'} &  {}={} s(a') - f \text{ and } \frac{d^2 J_{Bern}}{d a'^2} {}={} s(a')(1-s(a')).
\end{align}
Using matrix notation (where ${\bf s} \equiv s({\bf a})$), the gradients and Hessians are
\begin{align}
{\bf g}' {}={} & \boldsymbol\Psi^T \left(\boldsymbol\Psi {\bf s} - {\bf f} \right)\\
{\bf H}' {}={} & \boldsymbol\Psi^T \diag({\bf s})\diag(1-{\bf s}) \boldsymbol\Psi \simeq \diag\left( \boldsymbol\Psi^T \diag({\bf s}) (1-{\bf s}) \right)
\end{align}

\subsubsection{Categorical model}
For the categorical model with a softmax squashing function, the negative log-likelihood of a single voxel is
\begin{align}
J_{cat} & {}={} -\sum_{k=1}^K a'_{k} f_{k} + \log\left(\sum_{k=1}^K \exp a'_{k} \right)
\end{align}

This would use the gradients and Hessians
\begin{align}
\frac{d J_{cat}}{d a'_k} & {}={} s_k({\bf a}') - f_k \text{, where } s({\bf a}') = \frac{\exp {\bf a}')}{\sum_{k=1}^K \exp a'_k}\\
\frac{d^2 J_{cat}}{d a'_k a'_l} & {}={} s_k({\bf a}') (\delta_{jk} - s_j({\bf a}'))
\end{align}
where $\delta_{jk}$ is the Kronecker delta function.
Computation of the gradients and the approximation of the Hessian follow similar lines to those for the binary and Gaussian models. 

\subsection{Geodesic shooting}
The algorithm requires a means of computing diffeomorphic deformations from the initial velocities via a Geodesic shooting procedure.  Algorithm \ref{Alg:shoot} shows how this is achieved.
In the presented algorithm, $D\psi$ denotes the Jacobian tensor field of $\psi$, and $(D\psi)^T u$ indicates a pointwise multiplication with the transpose of the Jacobian.
$|D\psi|$ denotes the field of Jacobian determinants.
$L v$ in the continuous framework is equivalent to the matrix multiplication ${\bf L}^v {\bf v}$ in the discrete framework. 
The operation $L^g u$ denotes applying the inverse of $L$ to $u$, such that $L L^g u = u$.
In practice, this is a deconvolution, which is computed using fast Fourier transforms.
Much has already been written about the geodesic shooting procedure, so the reader is referred to \cite{miller2006geodesic,ashburner2011diffeomorphic} for further information.

\begin{algorithm}
\caption{Geodesic shooting via Euler integration}
\label{Alg:shoot}
\begin{algorithmic}
\Function{Shoot}{$v_0$}
    \State $u_0 \gets L v_0$ \Comment ${\bf L}^v {\bf v} \equiv L v$
    \State $\psi \gets id$
    \For{$t=1...T$}
        \State $u \gets \left|D\psi\right| (D\psi)^T u_0(\psi)$
        \State $v \gets L^g u$    \Comment Convolution using FFT
        \State $\psi \gets \psi(id - \tfrac{1}{T} v)$
    \EndFor
    \State \Return $\psi$ 
\EndFunction
\end{algorithmic}
\end{algorithm}

\subsection{Updating appearance basis functions ($\hat{\bf W}^a$)}
Appearance basis functions are optimised by
\begin{align}
\hat{\bf W}^a {}={} \argmin_{{\bf W}^a} \left(
    \tfrac{1}{2} \Tr\left((\lambda_1 N{\bf I} + \lambda_2 \hat{\bf Z}\hat{\bf Z}^T) ({\bf W}^a)^T {\bf L}^a {\bf W}^a \right)
    + \sum_{n=1}^N J({\bf f}_n, \hat{\bf z}_n, \hat{\boldsymbol\mu}, {\bf W}^a, \hat{\bf W}^v)
\right). \label{Eq:Wa_update}
\end{align}

The first step involves computing the gradients and Hessians, which can be done in a distributed way and is shown in Algorithm \ref{Alg:Appearance}.
Note that Algorithm \ref{Alg:Appearance} only shows the computation of gradients and Hessians for the Gaussian noise model, and that slight modifications are required when using other forms of appearance model.
Gradients and Hessians for updating these basis functions (${\bf W}^a$) are similar to those for the mean updates, except for weighting based on the current estimates of the latent variables.
Note that for this approach to work effectively, the rows of $\hat{\bf Z}$ should be orthogonal to each other, which is explained further in Section \ref{Sec:orthog}.
Note that only a single Gauss-Newton iteration is performed, so the objective function in Eqn. \ref{Eq:Wa_update} is not fully optimised, but merely improved over its previous value.

\begin{algorithm}
\caption{Computing gradients and Hessians for appearance}
\label{Alg:Appearance}
\begin{algorithmic}
\Function{AppearanceDerivatives}{${\bf F}, \hat{\bf Z}, \hat{\boldsymbol\mu}, \hat{\bf W}^a, \hat{\bf W}^v$}
    \For{$k=1...K^v$}
        \State ${\bf g}^a_k \gets {\bf 0}$, ${\bf H}^a_{kk} \gets {\bf 0}$
    \EndFor
    \For{$n=1...N$}
        \State ${\bf a}  \gets \hat{\boldsymbol\mu} + \hat{\bf W}^a \hat{\bf z}_n$
        \State $\boldsymbol\Psi \gets \text{Shoot}(\hat{\bf W}^v \hat{\bf z}_n)$
        \State ${\bf g}', {\bf H}' \gets \text{LikelihoodDerivatives}({\bf f}_n, {\bf a}, \boldsymbol\Psi)$
        \For{$k=1...K^v$}
            \State ${\bf g}^a_k \gets {\bf g}^a_k + \hat{z}_{kn} {\bf g}'$
            \State ${\bf H}^a_{kk} \gets {\bf H}^a_{kk} + \hat{z}_{kn}^2 {\bf H}'$
        \EndFor
    \EndFor
    \State \Return ${\bf G}^a, \mathcal{H}^a$
    \Comment Where ${\bf G}^a = \{{\bf g}^a_1, {\bf g}^a_2, \hdots, {\bf g}^a_K\}$
             and $\mathcal{H}^a = \{{\bf H}^a_{1,1}, {\bf H}^a_{2,2}, \hdots, {\bf H}^a_{K,K}\}$
\EndFunction
\end{algorithmic}
\end{algorithm}

\subsection{Updating shape basis functions ($\hat{\bf W}^v$)}
Shape basis functions are optimised by
\begin{align}
\hat{\bf W}^v {}={} \argmin_{{\bf W}^v} \left(
    \tfrac{1}{2} \Tr\left((\lambda_1 N {\bf I} + \lambda_2 \hat{\bf Z}\hat{\bf Z}^T) ({\bf W}^v)^T {\bf L}^v {\bf W}^v \right)
    + \sum_{n=1}^N J({\bf f}_n, \hat{\bf z}_n, \hat{\boldsymbol\mu}, \hat{\bf W}^a, {\bf W}^v)
\right). \label{Eq:Wv_update}
\end{align}

A single Gauss-Newton iteration is used to update the basis functions of the shape model (${\bf W}^v$), which is done in such a way that changes to ${\bf W}^v$ improve the objective function with respect to its previous value, rather than fully optimise Eqn. \ref{Eq:Wv_update}.
While most Gauss-Newton iterations improve the fit, a backtracking line search is included to ensure that they do not overshoot.
For simplicity, this aspect of the algorithm is not shown in Algorithm \ref{Alg:PG}.
As for updating ${\bf W}^a$, this requires the rows of $\hat{\bf Z}$ to be orthogonal to each other.

As for updating the appearance basis functions, computing the gradients and Hessians for the shape bases can be also done in a distributed way.
The strategy for computing gradients and Hessians is shown in Algorithm \ref{Alg:Shape}.

\begin{algorithm}
\caption{Computing gradients and Hessians for shape}
\label{Alg:Shape}
\begin{algorithmic}
\Function{ShapeDerivatives}{${\bf F}, \hat{\bf Z}, \hat{\boldsymbol\mu}, \hat{\bf W}^a, \hat{\bf W}^v$}
    \State \Comment Various settings (eg ${\bf L}^v$) are not passed as arguments
    \For{$k=1...K^v$}
        \State ${\bf g}^v_k \gets {\bf 0}$, ${\bf H}^v_{kk} \gets {\bf 0}$
    \EndFor
    \For{$n=1...N$}
        \State ${\bf a}  \gets \hat{\boldsymbol\mu} + \hat{\bf W}^a \hat{\bf z}_n$
        \State $\boldsymbol\Psi \gets \text{Shoot}(\hat{\bf W}^v \hat{\bf z}_n)$
        \State ${\bf g}', {\bf H}' \gets \text{LikelihoodDerivatives}({\bf f}_n, {\bf a}, \boldsymbol\Psi)$
        \State ${\bf D}  \gets \begin{bmatrix} \diag(\nabla_1 {\bf a}) & \diag(\nabla_2 {\bf a}) & \diag(\nabla_3 {\bf a}) \end{bmatrix}$
        \State ${\bf g}' \gets {\bf D}^T {\bf g}'$
        \State ${\bf H}' \gets {\bf D}^T {\bf H}' {\bf D}$
        \For{$k=1...K^v$} \Comment Update gradients and Hessians
            \State ${\bf g}^v_k \gets {\bf g}^v_k + \hat{z}_{kn} {\bf g}'_{n}$
            \State ${\bf H}^v_{kk} \gets {\bf H}^v_{kk} + \hat{z}_{kn}^2 {\bf H}'_{n}$
        \EndFor
    \EndFor
    \State \Return ${\bf G}^v, \mathcal{H}^v$
\EndFunction
\end{algorithmic}
\end{algorithm}

\subsection{Updating latent variables ($\hat{\bf z}_n$) \label{Sec:LatentUpdate}}
Within a distributed multi-centre privacy-preserving framework, the strategy would be to have the basis functions shared across sites, whereas the image data and latent variables would remain hidden within each of the sites.
All computations that use the image data and latent variables would be done at each site, with only aggregate data shared across them.
The modes of the latent variables are updated via a Gauss-Newton scheme, similar to that used by \cite{friston1995spatial,cootes2001active,cootes2001statistical}.

\begin{align}
\hat{\bf z}_n {}={} \argmin_{{\bf z}_n}\left(
J({\bf f}_n, {\bf z}_n, \boldsymbol\mu, \hat{\bf W}^a, \hat{\bf W}^v)
+ \tfrac{1}{2}{\bf z}_n^T \left(\lambda_1 \hat{\bf A} + \lambda_2 (\hat{\bf W}^a)^T {\bf L}^a \hat{\bf W}^a + \lambda_2 (\hat{\bf W}^v)^T {\bf L}^v \hat{\bf W}^v\right){\bf z}_n
\right)
\end{align}

\begin{algorithm}
\caption{Updating latent variables}
\label{Alg:Latent}
\begin{algorithmic}
\Function{UpdateLatentVariables}{${\bf F}, \hat{\bf Z}, \hat{\boldsymbol\mu}, \hat{\bf W}^a, \hat{\bf W}^v, {\bf A}$}
    \State ${\bf S} \gets {\bf 0}$
    \For{$n=1...N$}
        \State ${\bf a}  \gets \hat{\boldsymbol\mu} + \hat{\bf W}^a \hat{\bf z}_n$
        \State $\boldsymbol\Psi \gets \text{Shoot}(\hat{\bf W}^v \hat{\bf z}_n)$
        \State ${\bf g}', {\bf H}' \gets \text{LikelihoodDerivatives}({\bf f}_n, {\bf a}, \boldsymbol\Psi)$
        \State ${\bf D}  \gets \begin{pmatrix} \diag(\nabla_1 {\bf a}) & \diag(\nabla_2 {\bf a}) & \diag(\nabla_3 {\bf a}) \end{pmatrix}$
        \State ${\bf B} \gets {\bf D}^T {\bf W}^v + {\bf W}^a$
        \State ${\bf g} \gets {\bf B}^T {\bf g}'$
        \State ${\bf H} \gets {\bf B}^T {\bf H}' {\bf B}$
        \State $\hat{\bf z}_n \gets \hat{\bf z}_n - \left({\bf H} + {\bf A} \right)^{-1} \left({\bf g} + {\bf A} \hat{\bf z}_n \right)$
        \State ${\bf S} \gets {\bf S} + \left({\bf H} + {\bf A} \right)^{-1}$
    \EndFor
    \State ${\bf C}^z \gets \hat{\bf Z} \hat{\bf Z}^T$
    \State \Return $\hat{\bf Z}, {\bf S}, {\bf C}^z$
\EndFunction
\end{algorithmic}
\end{algorithm}

The inverse of the (approximate) Hessians allows a Gaussian approximation of the uncertainty with which the latent variables are updated to be computed (``Laplace approximation'').
This is the ${\bf S}$ matrix (returned by Algorithm \ref{Alg:Latent}), which is combined with $\hat{\bf Z} \hat{\bf Z}^T$ (returned as ${\bf C}^z$) and used to re-compute $\hat{\bf A}$.

\subsection{Expectation of the precision matrix ($\hat{\bf A}$) \label{Sec:A-update}}
This work uses a variational Bayesian approach for approximating the distribution of ${\bf A}$, which is a method described in more detail by textbooks, such as \cite{bishop2006pattern} or \cite{murphy2012machine}.
Briefly, it involves taking the joint probability of Eqn. \ref{Eqn:Objective}, discarding terms that do not involve ${\bf A}$, and substituting the expectations of the other parameters into the expression.
This leads to the following approximating distribution, which can be recognised as Wishart.
\begin{align}
\ln q({\bf A}) {}={} & \tfrac{1}{2}(N + \nu_0 - K - 1) \ln \det|{\bf A}| - \tfrac{1}{2}\Tr\left((\E[{\bf Z}{\bf Z}^T] + \boldsymbol\Lambda_0^{-1}) {\bf A}\right) + \text{const}\cr
 {}={} & \ln \mathcal{W}({\bf A} | \boldsymbol\Lambda, \nu)
\end{align}
where $\boldsymbol\Lambda = (\E[{\bf Z}{\bf Z}^T] + \boldsymbol\Lambda_0^{-1})^{-1}$  and $\nu = \nu_0 + N$.
In practice, $\E[{\bf Z}{\bf Z}^T]$ is approximated by ${\bf C}^z + {\bf S}$, described previously.

Other steps in the algorithm use the expectation of ${\bf A}$, which (see Appendix B of \cite{bishop2006pattern}) is
\begin{align}
\hat{\bf A} {}={} \E[{\bf A}] {}={} \nu \boldsymbol\Lambda. \label{Eqn:EA}
\end{align}

\subsection{Orthogonalisation \label{Sec:orthog}}
The strategy for updating $\hat{\bf W}^a$ and $\hat{\bf W}^v$ involves some approximations, which are needed in order to save memory and computation. This approximation is related to the Jacobi iterative method for determining the solutions to linear equations, which is only guaranteed to converge for diagonally dominant matrices.
Rather than work with the Hessian for the entire ${\bf W}$ matrix together, only the Hessians for each column of ${\bf W}$ are computed by Algorithms \ref{Alg:Appearance} and \ref{Alg:Shape}.
This corresponds with a block diagonal Hessian matrix for the entire ${\bf W}$, which has the form
\begin{align}
{\bf H} {}={} \begin{pmatrix}
{\bf H}_{11} & {\bf 0} & \hdots & {\bf 0}\cr
{\bf 0} & {\bf H}_{22} & \hdots & {\bf 0}\cr
\vdots       & \vdots & \ddots & \vdots\cr
{\bf 0} & {\bf 0} & \hdots & {\bf H}_{KK}
\end{pmatrix}.
\end{align}

More stable convergence can be achieved by transforming the basis functions and latent variables in order to minimise the amount of signal that would be in the off-diagonal blocks, thus increasing the diagonal dominance of the system of equations.
In situations where diagonal dominance is violated, convergence can still be achieved by decreasing the update step size.
This is analogous to using a weighted Jacobi iteration, where in practice the weights are found using a backtracking line-search.

Signal in the off-diagonal blocks is reduced by orthogonalising the rows of $\hat{\bf Z}$.
This is achieved by finding a transformation, ${\bf T}$, such that ${\bf T}\hat{\bf Z} ({\bf T} \hat{\bf Z})^T$ and  $(\hat{\bf W}^v{\bf T}^{-1})^T {\bf L}^v \hat{\bf W}^v{\bf T}^{-1}$ $+$ $(\hat{\bf W}^a{\bf T}^{-1})^T {\bf L}^a \hat{\bf W}^a{\bf T}^{-1}$ are both diagonal matrices.
Transformation ${\bf T}$ is derived from an eigendecomposition of the sufficient statistics, whereby the symmetric positive definite matrices are decomposed into diagonal (${\bf D}^z$ and ${\bf D}^w$) and orthonormal (${\bf V}^z$ and ${\bf V}^w$) matrices, such that
\begin{align}
{\bf V}^z {\bf D}^z ({\bf V}^z)^T {}={} & {\bf C}^z,\\
{\bf V}^w {\bf D}^w ({\bf V}^w)^T {}={} & {\bf C},
\end{align}
where ${\bf C}^z = \hat{\bf Z} \hat{\bf Z}^T$ and ${\bf C} = (\hat{\bf W}^v)^T {\bf L}^v \hat{\bf W}^v + (\hat{\bf W}^a)^T {\bf L}^a \hat{\bf W}^a$.

A further singular value decomposition is then used, giving
\begin{align}
{\bf U} {\bf D} {\bf V}^T {}={} ({\bf D}^w)^{\frac{1}{2}} ({\bf V}^w)^T {\bf V}^z ({\bf D}^z)^{\frac{1}{2}}.
\end{align}

The combination of various matrices is used to give an initial estimate of the transform
\begin{align}
{\bf T} {}={} {\bf D}{\bf V}^T ({\bf D}^z)^{-\frac{1}{2}} ({\bf V}^z)^T .\label{Eq:T}
\end{align}

The above ${\bf T}$ matrix could be used to render the matrices orthogonal, but their relative scalings would not be optimal. 
The remainder of the orthogonalisation procedure involves an iterative strategy similar to expectation maximisation, where the aim is to estimate some diagonal scaling matrix ${\bf Q}$ with which to multiply ${\bf T}$.
This matrix is parameterised by a set of parameters ${\bf q}$, such that
\begin{align}
{\bf Q} {}={} \diag(\exp {\bf q}).
\end{align}

The first step of the iterative scheme involves re-computing $\hat{\bf A}$, as described in Section \ref{Sec:A-update}, but incorporating the current estimates of ${\bf Q}{\bf T}$.
\begin{align}
\hat{\bf A} {}={} \nu \boldsymbol\Lambda =  (N + \nu_0)({\bf Q}{\bf T} ({\bf C}^z + {\bf S})({\bf Q}{\bf T})^T + \boldsymbol\Lambda_0^{-1})^{-1}.
\end{align}

The next step in the iterative scheme is to re-estimate ${\bf q}$, such that
\begin{align}
\hat{\bf q} {}={} \argmin_{\bf q} & ( \Tr\left( \diag(\exp (-{\bf q})) ({\bf T}^{-1})^T{\bf C}{\bf T}^{-1} \diag(\exp (-{\bf q})) \right) \cr
             + & \Tr\left(\diag(\exp {\bf q}){\bf T}{\bf C}^z {\bf T}^T\diag(\exp {\bf q}) \hat{\bf A} \right) ).
\end{align}
This can be achieved via a Gauss-Newton update, which uses first and second derivatives with respect to ${\bf q}$.
The overall strategy is illustrated in Algorithm \ref{Alg:Orthog}, which empirically is found to converge well.

\begin{algorithm}
\caption{Orthogonalising the variables}
\label{Alg:Orthog}
\begin{algorithmic}
\Function{OrthogonalisationMatrix}{${\bf C}, {\bf C}^z, {\bf S}, N$}
    \State ${\bf V}^z, {\bf D}^z \gets \text{eig}({\bf C}^z)$
    \State ${\bf V}^w, {\bf D}^w \gets \text{eig}({\bf C})$
    \State ${\bf U}, {\bf D}, {\bf V} \gets \text{svd}(({\bf D}^w)^{\frac{1}{2}} ({\bf V}^w)^T {\bf V}^z ({\bf D}^z)^{\frac{1}{2}})$
    \State ${\bf T} \gets {\bf D}{\bf V}^T ({\bf D}^z)^{-\frac{1}{2}} ({\bf V}^z)^T$
    \State ${\bf q} \gets {\bf 0}$
    \State ${\bf Q} \gets \diag(\exp {\bf q})$
    \Repeat
        \State $\hat{\bf A} \gets (N+\nu_0)({\bf Q}{\bf T}({\bf C}^z + {\bf S})({\bf Q}{\bf T})^T + \boldsymbol\Lambda_0^{-1})^{-1}$ \Comment See Eqn. \ref{Eqn:EA}.
        \State ${\bf R} \gets 2 \hat{\bf A} \odot ({\bf T}{\bf C}^z{\bf T}^T)^T$   \Comment ``$\odot$'' denotes a Hadamard product
        \State ${\bf g} \gets {\bf Q}{\bf R}\diag({\bf Q}) - 2 {\bf Q}^{-2}\diag(({\bf T}^{-1})^T {\bf C} {\bf T}^{-1})$ \Comment Gradient
        \State ${\bf H} \gets {\bf Q}{\bf R}{\bf Q} + \diag({\bf Q}{\bf R} \diag({\bf Q})) + 4 {\bf Q}^{-2} ({\bf T}^{-1})^T {\bf C} {\bf T}^{-1}$ \Comment Hessian
        \State ${\bf q} \gets {\bf q} - {\bf H}^{-1} {\bf g}$
        \State ${\bf Q} \gets \diag(\exp {\bf q})$
    \Until{Convergence}
    \State ${\bf T} \gets {\bf Q}{\bf T}$
    \State \Return ${\bf T}$
\EndFunction
\end{algorithmic}
\end{algorithm}

\bibliography{References}

\begin{thebibliography}{10}
\expandafter\ifx\csname url\endcsname\relax
  \def\url#1{\texttt{#1}}\fi
\expandafter\ifx\csname urlprefix\endcsname\relax\def\urlprefix{URL }\fi
\expandafter\ifx\csname href\endcsname\relax
  \def\href#1#2{#2} \def\path#1{#1}\fi

\bibitem{cootes1992active}
T.~F. Cootes, C.~J. Taylor, Active shape models -- `smart snakes', in: BMVC92,
  Springer, 1992, pp. 266--275.

\bibitem{cootes1995active}
T.~F. Cootes, C.~J. Taylor, D.~H. Cooper, J.~Graham, Active shape models --
  their training and application, Computer vision and image understanding
  61~(1) (1995) 38--59.

\bibitem{babalola2009evaluation}
K.~O. Babalola, B.~Patenaude, P.~Aljabar, J.~Schnabel, D.~Kennedy, W.~Crum,
  S.~Smith, T.~Cootes, M.~Jenkinson, D.~Rueckert, An evaluation of four
  automatic methods of segmenting the subcortical structures in the brain,
  Neuroimage 47~(4) (2009) 1435--1447.

\bibitem{patenaude2011bayesian}
B.~Patenaude, S.~M. Smith, D.~N. Kennedy, M.~Jenkinson, A {B}ayesian model of
  shape and appearance for subcortical brain segmentation, Neuroimage 56~(3)
  (2011) 907--922.

\bibitem{cootes2008diffeomorphic}
T.~Cootes, C.~Twining, K.~Babalola, C.~Taylor, Diffeomorphic statistical shape
  models, Image and Vision Computing 26~(3) (2008) 326--332.

\bibitem{rueckert2003automatic}
D.~Rueckert, A.~F. Frangi, J.~Schnabel, et~al., Automatic construction of 3-{D}
  statistical deformation models of the brain using nonrigid registration,
  Medical Imaging, IEEE Transactions on 22~(8) (2003) 1014--1025.

\bibitem{zhang2015bayesian}
M.~Zhang, P.~T. Fletcher, Bayesian principal geodesic analysis for estimating
  intrinsic diffeomorphic image variability, Medical Image Analysis.

\bibitem{adams2004geometric}
D.~Adams, F.~Rohlf, D.~Slice, Geometric morphometrics: {T}en years of progress
  following the `revolution', Italian Journal of Zoology 71~(1) (2004) 5--16.

\bibitem{zhang2017probabilistic}
M.~Zhang, W.~M. Wells, P.~Golland, Probabilistic modeling of anatomical
  variability using a low dimensional parameterization of diffeomorphisms,
  Medical Image Analysis.

\bibitem{cootes2001active}
T.~F. Cootes, G.~J. Edwards, C.~J. Taylor, Active appearance models, IEEE
  Transactions on Pattern Analysis \& Machine Intelligence 23~(6) (2001)
  681--685.

\bibitem{cootes2001statistical}
T.~F. Cootes, C.~J. Taylor, Statistical models of appearance for medical image
  analysis and computer vision, in: Medical Imaging 2001, International Society
  for Optics and Photonics, 2001, pp. 236--248.

\bibitem{belongie2002shape}
S.~Belongie, J.~Malik, J.~Puzicha, Shape matching and object recognition using
  shape contexts, Pattern Analysis and Machine Intelligence, IEEE Transactions
  on 24~(4) (2002) 509--522.

\bibitem{litjens2014evaluation}
G.~Litjens, R.~Toth, W.~van~de Ven, C.~Hoeks, S.~Kerkstra, B.~van Ginneken,
  G.~Vincent, G.~Guillard, N.~Birbeck, J.~Zhang, et~al., Evaluation of prostate
  segmentation algorithms for {MRI}: {T}he {PROMISE12} challenge, Medical image
  analysis 18~(2) (2014) 359--373.

\bibitem{cootes2010computing}
T.~F. Cootes, C.~J. Twining, V.~S. Petrovic, K.~O. Babalola, C.~J. Taylor,
  Computing accurate correspondences across groups of images, IEEE transactions
  on pattern analysis and machine intelligence 32~(11) (2010) 1994--2005.

\bibitem{alabort2014bayesian}
J.~Alabort-i Medina, S.~Zafeiriou, Bayesian active appearance models, in:
  Computer Vision and Pattern Recognition (CVPR), 2014 IEEE Conference on,
  IEEE, 2014, pp. 3438--3445.

\bibitem{lindner2015learning}
C.~Lindner, J.~Thomson, T.~F. Cootes, arcOGEN Consortium, et~al.,
  Learning-based shape model matching: {T}raining accurate models with minimal
  manual input, in: International Conference on Medical Image Computing and
  Computer-Assisted Intervention, Springer, 2015, pp. 580--587.

\bibitem{stern2016local}
D.~{\v{S}}tern, T.~Ebner, M.~Urschler, From local to global random regression
  forests: {E}xploring anatomical landmark localization, in: International
  Conference on Medical Image Computing and Computer-Assisted Intervention,
  Springer, 2016, pp. 221--229.

\bibitem{rasmussen2006gaussian}
C.~Rasmussen, C.~Williams, Gaussian processes for machine learning, Springer,
  2006.

\bibitem{lecun2015deep}
Y.~LeCun, Y.~Bengio, G.~Hinton, Deep learning, Nature 521~(7553) (2015)
  436--444.

\bibitem{mumford2002pattern}
D.~Mumford, Pattern theory: {T}he mathematics of perception, in: Proceedings of
  the International Congress of Mathematicians, Vol.~3, 2002.

\bibitem{goodfellow2014generative}
I.~Goodfellow, J.~Pouget-Abadie, M.~Mirza, B.~Xu, D.~Warde-Farley, S.~Ozair,
  A.~Courville, Y.~Bengio, Generative adversarial nets, in: Advances in Neural
  Information Processing Systems, 2014, pp. 2672--2680.

\bibitem{blaiotta2018generative}
C.~Blaiotta, P.~Freund, M.~J. Cardoso, J.~Ashburner, Generative diffeomorphic
  modelling of large mri data sets for probabilistic template construction,
  NeuroImage 166 (2018) 117--134.

\bibitem{beg2005computing}
M.~Beg, M.~Miller, A.~Trouv{\'e}, L.~Younes, Computing large deformation metric
  mappings via geodesic flows of diffeomorphisms, International Journal of
  Computer Vision 61~(2) (2005) 139--157.

\bibitem{miller2006geodesic}
M.~Miller, A.~Trouv{\'e}, L.~Younes, Geodesic shooting for computational
  anatomy, Journal of Mathematical Imaging and Vision 24~(2) (2006) 209--228.

\bibitem{bro1996fast}
M.~Bro-Nielsen, C.~Gramkow, Fast fluid registration of medical images, in:
  Visualization in Biomedical Computing, Springer, 1996, pp. 265--276.

\bibitem{christensen1996deformable}
G.~Christensen, R.~Rabbitt, M.~Miller, Deformable templates using large
  deformation kinematics, IEEE transactions on image processing 5~(10) (1996)
  1435--1447.

\bibitem{evans1995commentary}
A.~Evans, Commentary on ``spatial regulation and normalization of images'' by
  friston et al., Human Brain Mapping 3~(3) (1995) 254--256.

\bibitem{chang2017code}
L.~Chang, D.~Y. Tsao, The code for facial identity in the primate brain, Cell
  169~(6) (2017) 1013--1028.

\bibitem{lundqvist1998karolinska}
D.~Lundqvist, A.~Flykt, A.~{\"O}hman, The karolinska directed emotional faces
  (kdef), CD ROM from Department of Clinical Neuroscience, Psychology section,
  Karolinska Institutet (1998) 91--630.

\bibitem{friston2002classical}
K.~J. Friston, W.~Penny, C.~Phillips, S.~Kiebel, G.~Hinton, J.~Ashburner,
  Classical and {B}ayesian inference in neuroimaging: {T}heory, NeuroImage
  16~(2) (2002) 465--483.

\bibitem{datar2012mixed}
M.~Datar, P.~Muralidharan, A.~Kumar, S.~Gouttard, J.~Piven, G.~Gerig,
  R.~Whitaker, P.~T. Fletcher, Mixed-effects shape models for estimating
  longitudinal changes in anatomy, in: International Workshop on
  Spatio-temporal Image Analysis for Longitudinal and Time-Series Image Data,
  Springer, 2012, pp. 76--87.

\bibitem{allassonniere2015bayesian}
S.~Allassonni{\`e}re, S.~Durrleman, E.~Kuhn, Bayesian mixed effect atlas
  estimation with a diffeomorphic deformation model, SIAM Journal on Imaging
  Sciences 8~(3) (2015) 1367--1395.

\bibitem{groves2011linked}
A.~R. Groves, C.~F. Beckmann, S.~M. Smith, M.~W. Woolrich, Linked independent
  component analysis for multimodal data fusion, Neuroimage 54~(3) (2011)
  2198--2217.

\bibitem{worsley1995tests}
K.~J. Worsley, J.-B. Poline, A.~Vandal, K.~J. Friston, Tests for distributed,
  nonfocal brain activations, Neuroimage 2~(3) (1995) 183--194.

\bibitem{radford2015unsupervised}
A.~Radford, L.~Metz, S.~Chintala, Unsupervised representation learning with
  deep convolutional generative adversarial networks, arXiv preprint
  arXiv:1511.06434.

\bibitem{modat2014simulating}
M.~Modat, I.~J. Simpson, M.~J. Cardoso, D.~M. Cash, N.~Toussaint, N.~C. Fox,
  S.~Ourselin, Simulating neurodegeneration through longitudinal population
  analysis of structural and diffusion weighted {MRI} data, in: International
  Conference on Medical Image Computing and Computer-Assisted Intervention,
  Springer, 2014, pp. 57--64.

\bibitem{lecun1998gradient}
Y.~LeCun, L.~Bottou, Y.~Bengio, P.~Haffner, Gradient-based learning applied to
  document recognition, Proceedings of the IEEE 86~(11) (1998) 2278--2324.

\bibitem{mackay2003information}
D.~J. MacKay, Information theory, inference and learning algorithms, Cambridge
  university press, 2003.

\bibitem{hinton2011transforming}
G.~E. Hinton, A.~Krizhevsky, S.~D. Wang, Transforming auto-encoders, in:
  International Conference on Artificial Neural Networks, Springer, 2011, pp.
  44--51.

\bibitem{taigman2014deepface}
Y.~Taigman, M.~Yang, M.~Ranzato, L.~Wolf, Deepface: {C}losing the gap to
  human-level performance in face verification, in: Proceedings of the IEEE
  conference on computer vision and pattern recognition, 2014, pp. 1701--1708.

\bibitem{jaderberg2015spatial}
M.~Jaderberg, K.~Simonyan, A.~Zisserman, et~al., Spatial transformer networks,
  in: Advances in Neural Information Processing Systems, 2015, pp. 2017--2025.

\bibitem{revow1996using}
M.~Revow, C.~K. Williams, G.~E. Hinton, Using generative models for handwritten
  digit recognition, IEEE transactions on pattern analysis and machine
  intelligence 18~(6) (1996) 592--606.

\bibitem{bishop2006pattern}
C.~Bishop, et~al., Pattern recognition and machine learning, Springer New
  York:, 2006.

\bibitem{murphy2012machine}
K.~P. Murphy, Machine learning: {A} probabilistic perspective, MIT press, 2012.

\bibitem{lee2015deeply}
C.-Y. Lee, S.~Xie, P.~Gallagher, Z.~Zhang, Z.~Tu, Deeply-supervised nets, in:
  Artificial Intelligence and Statistics, 2015, pp. 562--570.

\bibitem{bruna2013invariant}
J.~Bruna, S.~Mallat, Invariant scattering convolution networks, IEEE
  transactions on pattern analysis and machine intelligence 35~(8) (2013)
  1872--1886.

\bibitem{cootes1999mixture}
T.~F. Cootes, C.~J. Taylor, A mixture model for representing shape variation,
  Image and Vision Computing 17~(8) (1999) 567--573.

\bibitem{van2010capturing}
L.~Van Der~Maaten, E.~Hendriks, Capturing appearance variation in active
  appearance models, in: Computer Vision and Pattern Recognition Workshops
  (CVPRW), 2010 IEEE Computer Society Conference on, IEEE, 2010, pp. 34--41.

\bibitem{gooya2015joint}
A.~Gooya, K.~Lekadir, X.~Alba, A.~Swift, J.~M. Wild, A.~F. Frangi, Joint
  clustering and component analysis of correspondenceless point sets:
  {A}pplication to cardiac statistical modeling, in: Information Processing in
  Medical Imaging: 24th International Conference, IPMI 2015, Sabhal Mor Ostaig,
  Isle of Skye, UK, June 28-July 3, 2015, Proceedings, Vol. 9123, Springer,
  2015, p.~98.

\bibitem{le2017spectral}
L.~Le~Folgoc, A.~V. Nori, A.~Criminisi, Spectral kernels for probabilistic
  analysis and clustering of shapes, in: International Conference on
  Information Processing in Medical Imaging, Springer, 2017, pp. 67--79.

\bibitem{marcus2010open}
D.~Marcus, A.~Fotenos, J.~Csernansky, J.~Morris, R.~Buckner, Open access series
  of imaging studies: {L}ongitudinal {MRI} data in nondemented and demented
  older adults, Journal of cognitive neuroscience 22~(12) (2010) 2677--2684.

\bibitem{ashburner2005unified}
J.~Ashburner, K.~Friston, Unified segmentation, Neuroimage 26~(3) (2005)
  839--851.

\bibitem{weiskopf2011unified}
N.~Weiskopf, A.~Lutti, G.~Helms, M.~Novak, J.~Ashburner, C.~Hutton, Unified
  segmentation based correction of {R1} brain maps for {RF} transmit field
  inhomogeneities ({UNICORT}), Neuroimage 54~(3) (2011) 2116--2124.

\bibitem{malone2015accurate}
I.~B. Malone, K.~K. Leung, S.~Clegg, J.~Barnes, J.~L. Whitwell, J.~Ashburner,
  N.~C. Fox, G.~R. Ridgway, Accurate automatic estimation of total intracranial
  volume: {A} nuisance variable with less nuisance, Neuroimage 104 (2015)
  366--372.

\bibitem{gower1975generalized}
J.~C. Gower, Generalized procrustes analysis, Psychometrika 40~(1) (1975)
  33--51.

\bibitem{ghiassian2016using}
S.~Ghiassian, R.~Greiner, P.~Jin, M.~R. Brown, Using functional or structural
  magnetic resonance images and personal characteristic data to identify {ADHD}
  and autism, PloS one 11~(12) (2016) e0166934.

\bibitem{cabral2016classifying}
C.~Cabral, L.~Kambeitz-Ilankovic, J.~Kambeitz, V.~D. Calhoun, D.~B. Dwyer,
  S.~Von~Saldern, M.~F. Urquijo, P.~Falkai, N.~Koutsouleris, Classifying
  schizophrenia using multimodal multivariate pattern recognition analysis:
  {E}valuating the impact of individual clinical profiles on the
  neurodiagnostic performance, Schizophrenia bulletin 42~(suppl\_1) (2016)
  S110--S117.

\bibitem{nieuwenhuis2012classification}
M.~Nieuwenhuis, N.~E. van Haren, H.~E.~H. Pol, W.~Cahn, R.~S. Kahn, H.~G.
  Schnack, Classification of schizophrenia patients and healthy controls from
  structural {MRI} scans in two large independent samples, Neuroimage 61~(3)
  (2012) 606--612.

\bibitem{monte2018comparison}
G.~C. Mont{\'e}-Rubio, C.~Falc{\'o}n, E.~Pomarol-Clotet, J.~Ashburner, A
  comparison of various mri feature types for characterizing whole brain
  anatomical differences using linear pattern recognition methods, NeuroImage.

\bibitem{ashburner2012symmetric}
J.~Ashburner, G.~R. Ridgway, Symmetric diffeomorphic modeling of longitudinal
  structural {MRI}, Frontiers in neuroscience 6.

\bibitem{tipping1999probabilistic_pri}
M.~Tipping, C.~Bishop, Probabilistic principal component analysis, Journal of
  the Royal Statistical Society. Series B, Statistical Methodology (1999)
  611--622.

\bibitem{allassonniere2010construction}
S.~Allassonni{\`e}re, E.~Kuhn, A.~Trouv{\'e}, et~al., Construction of
  {B}ayesian deformable models via a stochastic approximation algorithm: {A}
  convergence study, Bernoulli 16~(3) (2010) 641--678.

\bibitem{zhang2013probabilistic}
M.~Zhang, T.~Fletcher, Probabilistic principal geodesic analysis, in: Advances
  in Neural Information Processing Systems, 2013, pp. 1178--1186.

\bibitem{tipping1999probabilistic_vis}
M.~E. Tipping, Probabilistic visualisation of high-dimensional binary data,
  Advances in neural information processing systems (1999) 592--598.

\bibitem{elhabian2017label}
S.~Y. Elhabian, R.~T. Whitaker, From label maps to generative shape models: {A}
  variational {B}ayesian learning approach, in: International Conference on
  Information Processing in Medical Imaging, Springer, 2017, pp. 93--105.

\bibitem{bhatia2007groupwise}
K.~K. Bhatia, P.~Aljabar, J.~P. Boardman, L.~Srinivasan, M.~Murgasova, S.~J.
  Counsell, M.~A. Rutherford, J.~V. Hajnal, A.~D. Edwards, D.~Rueckert,
  Groupwise combined segmentation and registration for atlas construction, in:
  International Conference on Medical Image Computing and Computer-Assisted
  Intervention, Springer, 2007, pp. 532--540.

\bibitem{purwani2014ensemble}
S.~Purwani, T.~F. Cootes, C.~J. Twining, Ensemble registration: {C}ombining
  groupwise registration and segmentation., in: MIUA, 2014, pp. 105--110.

\bibitem{balbastre2018diffeomorphic}
Y.~Balbastre, M.~Brudfors, K.~Bronik, J.~Ashburner, Diffeomorphic brain shape
  modelling using gauss-newton optimisation, arXiv preprint arXiv:1806.07109.

\bibitem{allassonniere2007towards}
S.~Allassonni{\`e}re, Y.~Amit, A.~Trouve, Towards a coherent statistical
  framework for dense deformable template estimation, Journal of the Royal
  Statistical Society: Series B (Statistical Methodology) 69~(1) (2007) 3--29.

\bibitem{sang2012full}
H.~Sang, J.~Z. Huang, A full scale approximation of covariance functions for
  large spatial data sets, Journal of the Royal Statistical Society: Series B
  (Statistical Methodology) 74~(1) (2012) 111--132.

\bibitem{niethammer2011geodesic}
M.~Niethammer, Y.~Huang, F.~Vialard, Geodesic regression for image time-series,
  in: Medical Image Computing and Computer-Assisted Intervention--MICCAI 2011,
  Springer, 2011, pp. 655--662.

\bibitem{fletcher2013geodesic}
P.~T. Fletcher, Geodesic regression and the theory of least squares on
  {R}iemannian manifolds, International journal of computer vision 105~(2)
  (2013) 171--185.

\bibitem{bach2005probabilistic}
F.~Bach, M.~Jordan, A probabilistic interpretation of canonical correlation
  analysis, Dept. Statist., Univ. California, Berkeley, CA, Tech. Rep 688.

\bibitem{klami2013bayesian}
A.~Klami, S.~Virtanen, S.~Kaski, Bayesian canonical correlation analysis,
  Journal of Machine Learning Research 14~(Apr) (2013) 965--1003.

\bibitem{ashburner2007fast}
J.~Ashburner, A fast diffeomorphic image registration algorithm, Neuroimage
  38~(1) (2007) 95--113.

\bibitem{ashburner2011diffeomorphic}
J.~Ashburner, K.~Friston, Diffeomorphic registration using geodesic shooting
  and {G}auss-{N}ewton optimisation, Neuroimage 55 (2011) 954--967.

\bibitem{friston1995spatial}
K.~Friston, J.~Ashburner, C.~D. Frith, J.-B. Poline, J.~D. Heather, R.~S.
  Frackowiak, et~al., Spatial registration and normalization of images, Human
  brain mapping 3~(3) (1995) 165--189.

\end{thebibliography}
\end{document}